\documentclass{article}

\usepackage[main, final]{neurips_2026}
\usepackage[utf8]{inputenc} % allow utf-8 input
\usepackage[T1]{fontenc}    % use 8-bit T1 fonts
\usepackage{hyperref}       % hyperlinks
\usepackage{url}            % simple URL typesetting
\usepackage{booktabs}       % professional-quality tables
\usepackage{amsfonts}       % blackboard math symbols
\usepackage{amsmath}
\usepackage{amssymb}
\usepackage{amsthm}
\usepackage{nicefrac}       % compact symbols for 1/2, etc.
\usepackage{microtype}      % microtypography
\usepackage{xcolor}         % colors
\usepackage{graphicx} % Required for inserting images

\usepackage{algorithm}
\usepackage[noend]{algpseudocode}
\usepackage{xcolor}
\usepackage{subcaption}
\usepackage{cleveref}
\usepackage{float}
\usepackage{array}
\usepackage{multirow}

\title{AERO: Adaptive Ensemble-Disagreement Routing for Oracle Feedback in Sample-Efficient Online RLHF}
\author{
  Shrey Singhal \\
  RMIT University \\
  Melbourne, VIC, Australia \\
  \texttt{shrey.singhal@student.rmit.edu.au}
  \And
  Nannan Huang \\
  RMIT University \\
  Melbourne, VIC, Australia \\
  \texttt{amber.huang@student.rmit.edu.au}
  \And
  Haytham M. Fayek \\
  RMIT University \\
  Melbourne, VIC, Australia \\
  \texttt{haytham.fayek@ieee.org}
}
\begin{document}

\maketitle

\begin{abstract}
Aligning Large Language Models (LLMs) with human preferences via Reinforcement Learning from Human Feedback (RLHF) has driven substantial improvements across a wide range of tasks, but the cost of acquiring preference labels remains the dominant bottleneck in the alignment pipeline. While many recent alignment methods improve sample efficiency through active exploration over candidate responses, they typically query the oracle uniformly across contexts, even for contexts where the learned reward estimates are already reliable enough to provide synthetic labels. We propose \textbf{A}daptive \textbf{E}nsemble-disagreement \textbf{R}outing for \textbf{O}racle feedback (AERO), a fully online RLHF algorithm built on a Dyna-style model-based RL architecture, in which an ensemble-based Epistemic Reward Model (ERM) serves both as a reward posterior for exploration and as a learned preference-feedback model for synthetic labeling. AERO uses ensemble vote entropy, a Query-by-Committee disagreement measure, combined with an adaptive thresholding mechanism to decide which contexts require oracle feedback and which can instead receive ERM-based synthetic preference labels, concentrating expensive labels where they are most informative. Empirical results across different model families and scales show that AERO achieves highly competitive alignment performance while reducing the required oracle-query budget compared to online DPO and strong active-exploration baselines.

\end{abstract}

\section{Introduction}

Reinforcement Learning from Human Feedback (RLHF) has emerged as a key paradigm for aligning Large Language Models (LLMs) with human preferences and values~\citep{christiano2017deep, bai2022training, ouyang2022training}. The standard alignment pipeline typically starts with a pretrained auto-regressive LLM~\citep{ouyang2022training}, which is supervised fine-tuned on high-quality demonstrations, then used to generate candidate responses that are ranked by an oracle to train a parameterized reward model, and finally optimized using a reinforcement learning algorithm such as Proximal Policy Optimization (PPO)~\citep{schulman2017proximal}. More recently, Direct Preference Optimization (DPO)~\citep{rafailov2023direct} has been proposed as an alternative that bypasses explicit reward modeling; instead, it learns directly from pairwise data by mapping the implicit reward function to the optimal policy via a binary cross-entropy objective that favors the chosen response over the rejected one. Despite algorithmic advancements such as DPO, the overarching alignment process remains bottlenecked by data inefficiency. 

Traditionally, RLHF pipelines collect preference data by sampling contexts uniformly at random from a given corpus, and improvements are often pursued by simply scaling annotation volume. However, recent data-centric perspectives on LLM post-training emphasize that such scaling yields diminishing returns, motivating more selective and value-aware data acquisition~\citep{luo2025survey}. Recent theoretical analysis by~\citet{das2025active} sharpens this in the RLHF setting, showing that under practical labeling budgets, uniform sampling of contexts forces the aligned policy to suffer a constant sub-optimality gap. At the same time, recent fully-online RLHF methods have improved sample efficiency by incorporating active exploration directly into the interaction loop, typically instantiating exploration at the level of action selection or pairwise comparison conditioned on a context or through optimistic policy objectives (e.g.,~\citep{dwaracherla2024efficient, liu2024sample, zhang2024self, asghari2026efficient, bai2025online}). 
However, both threads share an implicit assumption: that every context is equally deserving of oracle supervision. In practice, the reward model's epistemic uncertainty varies sharply across prompts, and uniformly labeling those where the model already exhibits strong consensus wastes budget on non-informative labels. Motivated by this gap, we investigate whether adaptively allocating oracle supervision based on context-level epistemic uncertainty can improve sample efficiency in fully online RLHF.

\begin{figure}
  \centering
  \includegraphics[width=\textwidth]{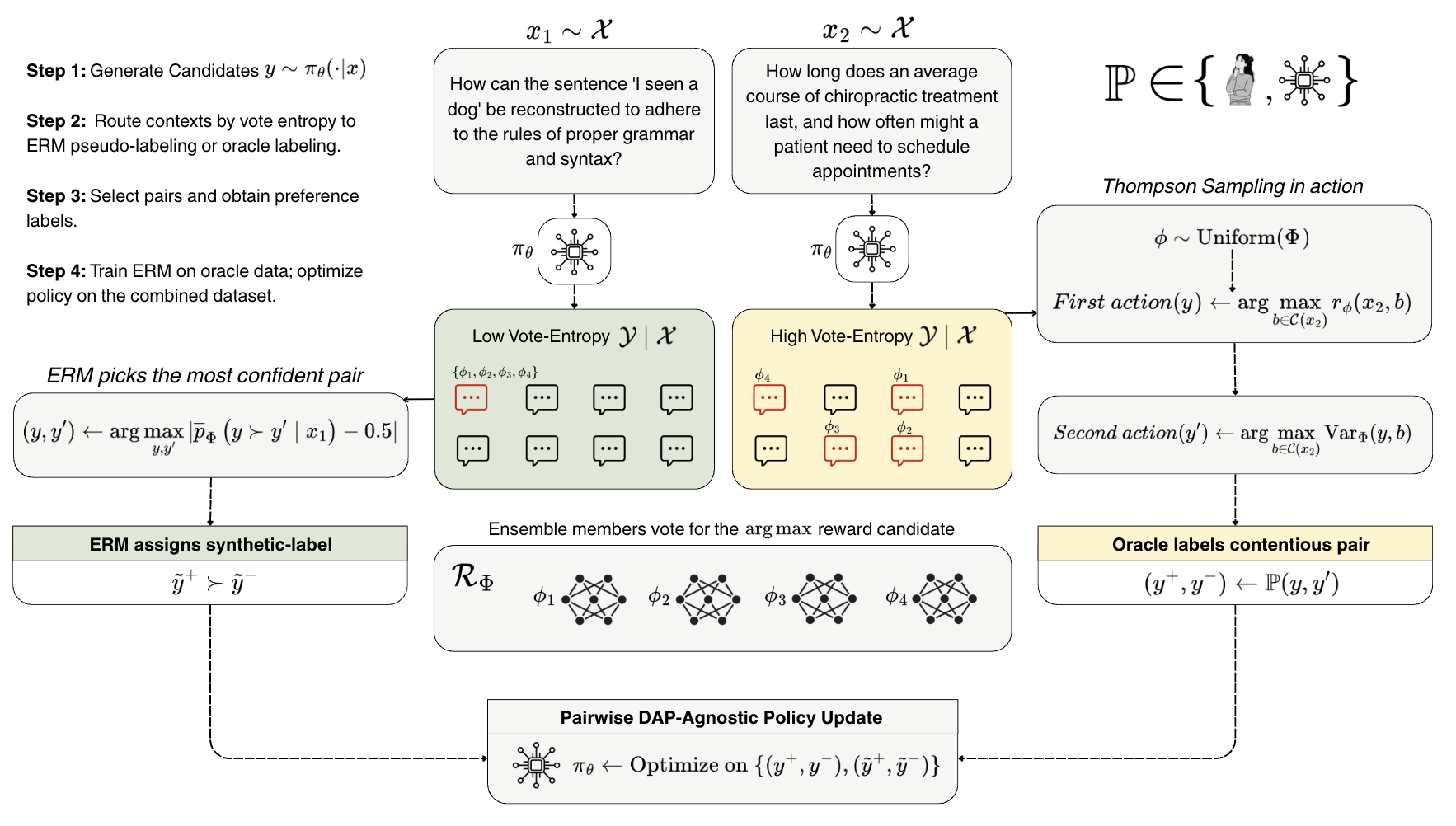}
  \caption{\textbf{Overview of AERO.} Contexts with high ensemble disagreement are routed to the oracle, while the rest receive synthetic labels from the ERM. $\mathbb{P}$ denotes the Preference Oracle, $\mathcal{R}_\Phi$ the ERM, $\pi_\theta$ the policy. Thompson Sampling explores the exact disagreement measured by vote entropy. High vote entropy ensures $y$ is drawn from severe disagreement.}
  \label{fig:aero_architecture}
\end{figure}

In this paper, we propose a novel sample-efficient RLHF algorithm named \textbf{A}daptive \textbf{E}nsemble-Disagreement \textbf{R}outing for \textbf{O}racle Feedback (AERO) (see Figure~\ref{fig:aero_architecture}). We adopt a query-by-committee (QBC)~\citep{seung1992query} inspired acquisition strategy to identify which examples warrant expensive oracle labeling, since QBC is designed to select informative examples from an unlabeled stream and is classically motivated by strong sample-efficiency guarantees in its theoretical setting~\citep{freund1997selective}. To instantiate this as a concrete routing score, we use vote entropy, following the extension of QBC to probabilistic classifiers by~\citet{argamon1999committee}, where informativeness is measured by committee disagreement, and vote entropy is interpreted as uncertainty over the predictions of plausible models rather than the entropy of a single model's predictive distribution. Inspired by SEA~\citep{liu2024sample}, we use an Epistemic Reward Model (ERM) as both an uncertainty-aware approximation to the reward posterior and a learned preference-feedback model for synthetic labeling, echoing Dyna-style model-based RL, where learned models support additional policy-improvement updates~\citep{sutton1990integrated,janner2019trust}. Implemented as a deep ensemble, the ERM enables posterior sampling for exploration and supplies synthetic preference labels for lower-disagreement contexts routed away from the oracle. %In contrast to prior active preference methods that improve label efficiency by selecting a smaller, high-value subset from a larger candidate batch or pool (e.g., \citep{muldrew2024active, das2025active}), our method operates in a fully online regime and instead improves sample efficiency through adaptive within-batch routing of contexts between expensive oracle supervision and ERM-generated synthetic preference labels.%

Our main contributions are as follows:  
%Notably, while AERO shares \citet{liu2024sample} response selection strategy, it replaces SEA's fixed-probability oracle assignment with adaptive prompt level routing based on vote entropy, making oracle allocation adaptive to prompt informativeness rather than random.% 
(i) We provide a principled QBC-inspired instantiation for online preference learning, where vote entropy over a deep ensemble ERM serves as a theoretically grounded measure of ensemble disagreement. (ii) We propose AERO, a fully online RLHF algorithm that uses this signal with adaptive batch-level thresholding to allocate oracle feedback only to high-disagreement contexts. (iii) We evaluate AERO against strong online RLHF baselines across two model families and three scales (1B to 8B), demonstrating improved sample efficiency on win-rate against the initial policy and competitive instruction-following performance on AlpacaEval 2.0.\footnote{Code is available at \url{https://github.com/Shrey-Singhal/AERO}.}

\section{Preliminaries}
\label{gen_inst}

\paragraph{Problem Statement.} We formalize the problem of sample-efficient online preference alignment as a budgeted Contextual Dueling Bandit problem (CDB)~\citep{dudik2015contextual, saha2021optimal}, where an agent must learn an optimal policy through pairwise preference feedback under a constrained oracle-query budget. Let $\mathcal{X}$ denote the context space and $\mathcal{Y}$ the action space. The parameterized language policy $\pi_\theta$ associates each context $x \in \mathcal{X}$ with a probability distribution $\pi(\cdot|x) \in \Delta(\mathcal{Y})$ over the action space. At each time step $t \in [T]$, where $T$ denotes the total number of interaction rounds, given $x_t \sim \mathcal{X}$, we draw a candidate set $\mathcal{C}_t (x_t)$ from $\pi_{\theta_t}$, then select a dueling pair $(y_t, y'_t) \in \mathcal{C}_t (x_t)$ for preference comparison. We model preferences via the Bradley-Terry model~\citep{bradley1952rank}, which assumes a latent reward function $r^\star(x, y)$ such that $p(y_t \succ y'_t \ | \ x_t) = \sigma (r^\star(x_t, y_t) - r^\star(x_t, y'_t))$ where $\sigma$ is the sigmoid function. We measure sample efficiency as cumulative query consumption $B$ throughout training. We seek comparable or improved alignment at reduced $B$.

\paragraph{Vote Entropy for Committee-based Acquisition.} Following~\citet{argamon1999committee}, we quantify ensemble disagreement using vote entropy, which is the Shannon entropy of the empirical distribution of ensemble votes over candidate actions. For a context $x$, let $V(j;x)$ denote the number of ensemble members that vote for action $j$ among $N$ candidates, and let $\hat{p_j}(x) = V(j;x) / K$, where K is the committee size. We use the unnormalized vote entropy:
\begin{equation}
    VE(x) = - \sum^N_{j=1}\hat{p}_j(x)log \ \hat{p}_j(x).
\label{3}
\end{equation}
In our implementation, each ensemble member votes for the highest-reward candidate completion among the $N$ completions associated with a context.

\paragraph{Reward Modeling from Preference Data.} Given the oracle-labeled pairwise preference dataset collected up to round $t$, $\mathcal{D}_t = \{ (x_i, y^+_i, y^-_i)\}^N_{i=1} $,
where $y^+_i$ is preferred over $y^-_i$ for context $x_i$, we learn a parameterized reward model $r_\phi(x, y)$ by maximum likelihood under the Bradley-Terry model. This yields the pairwise negative log-likelihood loss:
\begin{equation}
    \mathcal{L}_r(\phi | \mathcal{D}_t) = -\mathbb{E}_{(x, y^+, y^-) \sim \mathcal{D}_t} \ \left[\text{log} \ \sigma(r_\phi(x, y^+) - r_\phi(x, y^-))\right].
\label{1}
\end{equation}
% The equation above is just the loss function used to train the reward model - in our case, it's the ensemble reward model%
This objective trains $r_\phi$ to assign a higher reward to preferred candidate actions, which yields an approximation to the latent ground truth reward $r^\star$.

\paragraph{Epistemic Reward Model.} We instantiate the reward model as a deep ensemble, %\mathcal{R}_\Phi\{r_{\Phi^k_t}\}^K_{k=1}$ %
inspired by prior work~\citep{liu2024sample, dwaracherla2024efficient}, to maintain the reward posterior $p(r|\mathcal{D})$ for active exploration, due to its ability to model epistemic uncertainty~\citep{lakshminarayanan2017simple}. Each ensemble member is trained independently and online on oracle-labeled data. Following this formulation, we train the ensemble via the regularized negative log-likelihood loss:
\begin{equation}
    \mathcal{L}_{ERM}(\Phi^t | \mathcal{D}_t) = \sum^{K}_{k=1} \ \left[\mathcal{L}_r(\phi^t_k \ | \ \mathcal{D}_t) + \lambda || \phi^t_k - \phi^0_k ||^2\right],
\label{2}
\end{equation}
where $\mathcal{L}_r$ is defined in~\eqref{1}, $\Phi^t = \{\phi^t_k \}^K_{k=1}$ contains the weights of the $K$-sized ensemble, and the regularization term $\lambda$ encourages diversity across ensemble members by anchoring each head to its own initial weights $\phi^0_k$. Following~\citet{liu2024sample}, we refer to this ensemble as the Epistemic Reward Model (ERM), denoted as $\mathcal{R}_\Phi$.

\section{Adaptive Ensemble-disagreement Routing for Oracle Feedback (AERO)}
\label{sec:aero}
AERO allocates oracle feedback to contexts with high ERM disagreement and uses ERM synthetic labels for lower-disagreement contexts. At each round, the algorithm first routes contexts using vote entropy and an adaptive threshold, then constructs preference pairs according to the assigned feedback source. Figure~\ref{fig:aero_architecture} illustrates this pipeline, and Algorithm~\ref{alg:aero} details the full procedure.

\subsection{Context Selection and Adaptive Routing}

At each round $t$, given a batch of $M$ contexts $\{x_i\}^M_{i=1}$, we score each context using $VE(x)$~\eqref{3} to measure the ensemble disagreement. This aligns with our Thompson-sampling strategy for picking the first action where we draw a plausible reward model from the ERM and select its highest-reward candidate, hence, $VE(x)$ quantifies ensemble disagreement about which action would be selected. Following the QBC principle \citep{seung1992query, freund1997selective, settles2009active} that committee disagreement identifies examples where oracle labels are expected to be most useful, we route high $VE(x)$ contexts to the oracle and low $VE(x)$ contexts to ERM synthetic labeling.

\paragraph{Adaptive Thresholding.} We use Otsu's method \citep{otsu1979threshold} to split the batch-level vote-entropy scores into high- and low-disagreement groups. Although introduced for image thresholding, Otsu-style thresholding has also been used to partition unlabeled examples by OOD-related scores in open-set and safe semi-supervised learning~\citep{yu2020multi, ma2023rethinking}. Given the batch of sorted vote entropy scores $S_t = VE\{(x_i)\}^M_{i=1}$, we evaluate each split index $\tau$ and choose the one maximizing between group variance:
\begin{equation}
    \tau^* = \arg\max_\tau w_0(\tau)w_1(\tau)(u_0(\tau) - u_1(\tau))^2,
    \label{4}
\end{equation}
where $w_0(\tau)$ and $w_1(\tau)$ denote the fractions of scores below and above $\tau$, and $u_0(\tau)$ and $u_1(\tau)$ are the corresponding means of both groups. This yields a data-dependent routing threshold that adapts to the uncertainty profile of each online batch without extensive hyperparameter search.

\subsection{Dueling  Pair Selection}
\label{sec:pair selectionn}
Pair selection is supervision-dependent as oracle-labeled pairs prioritize information gain from expensive feedback whereas ERM-labeled pairs prioritize synthetic-label reliability. We therefore use branch-specific selection rules with safeguards against intrinsically ambiguous comparisons. For contexts routed to oracle, the goal is to select the pair that maximizes information gain. Hence, we select the first action (the anchor)  via Thompson sampling~\citep{thompson1933likelihood}, following recent TS-based exploration methods for LLM alignment~\citep{dwaracherla2024efficient, liu2024sample}. The second action (the challenger) is chosen as the candidate whose preference against the anchor has the highest ensemble variance. This follows the best-arm identification (BAI) design of~\citet{liu2024sample}, using reliable oracle feedback to probe epistemically uncertain comparisons. For ERM-routed contexts, pair selection instead prioritizes label reliability by choosing the pair among all candidate pairs $(a, b) \in \mathcal{C}_t(x)$ that has the highest ensemble-mean confidence, measured as $|\,\bar{p}(a \succ b \mid x) - 0.5\,|$, and assigning the synthetic label from the ensemble-mean preference probability.

\paragraph{Aleatoric-Uncertainty Filtering.} To avoid training on intrinsically ambiguous comparisons, we apply a pair-level filter before final pair selection. Following uncertainty decomposition arguments discussed in Appendix~\ref{Motivation for Aleatoric Filtering}, we use the mean binary entropy across ERM heads as a proxy for measuring aleatoric uncertainty for a candidate pair $(a,b)$ under context $x$:
\begin{equation}
    A (x, a, b) = \frac{1}{K}\sum^K_{k=1}\mathcal{H}(p_k(a \succ b \ | \ x)).
\label{6}
\end{equation}
Pairs with $A(x,a,b) > \alpha$  are excluded from training. We set $\alpha = 0.6$, since binary entropy under the natural logarithm is maximized at $\approx 0.693$ when $p = 0.5$, and $\mathcal{H}(p) = 0.6$ corresponds to $p \in [0.29, 0.71]$, meaning pairs above this threshold exhibit near-uniform preference probability and are therefore intrinsically ambiguous rather than merely uncertain. For oracle-routed contexts, we filter challengers against the Thompson-sampled anchor before selecting the highest-variance admissible challenger; for ERM-routed contexts, we filter all candidate pairs before selecting the most confident remaining pair. Contexts with no admissible pair are dropped.

\begin{algorithm}[t]
\caption{AERO procedure}
\label{alg:aero}
\textbf{Input:} Context distribution $c_\mathcal{X}$, preference oracle $\mathbb{P}$, loss function $F$

Initialize policy $\pi_{\theta_0} \leftarrow \pi_{SFT}$, and ERM weights $\Phi^0 = \{\phi^0_k\}^K_{k=1}$ randomly 
\begin{algorithmic}[1]
    \For{$t = 1, \ldots, T$}
        \State Sample a batch of contexts $\mathcal{X}_t = \{x_i\}^M_{i=1}$ from  $c_\mathcal{X}$
        \State For each context, sample $N$ candidate completions from $\pi_{\theta_{t-1}}(\cdot |x_i)$ to form $\mathcal{C}_t(x_i)$
        \State Compute the batch of vote-entropy scores $\mathcal{S}_t=\{VE(x_i)\}_{i=1}^M$ using Eq.~\eqref{3}
        \State Compute the routing threshold $\tau^\star$ from $\mathcal{S}_t$ using Eq.~\eqref{4}

        \State Initialize $D_t^O \leftarrow \emptyset,\; D_t^S \leftarrow \emptyset$
        \For{$x_i$ in $\mathcal{X}_t$}
            \Statex \hspace{2.7em} {\color{gray}\itshape // Route to Oracle branch}
            \If{$VE(x_i) > \tau_t^\star$}
                \Statex \hspace{4.2em} {\color{gray}\itshape // Select Oracle anchor}
                \State Sample $\phi \sim \mathrm{Uniform}(\Phi^{t-1})$ and set $y_i \leftarrow \arg\max_{b \in \mathcal{C}_t(x_i)} r_\phi(x_i,b)$
                \State $\mathcal{B}_i \leftarrow \{\, b \in \mathcal{C}_t(x_i)\setminus\{y_i\} \;:\; A(x_i,y_i,b) \le \alpha \,\}$, where $A$ is defined in Eq.~\eqref{6}
                \If{$\mathcal{B}_i \neq \emptyset$}
                    \Statex \hspace{5.7em} {\color{gray}\itshape // Select Oracle challenger}
                    \State $y_i' \leftarrow \arg\max_{b \in \mathcal{B}_i}
                    \mathrm{Var}_{k=1}^K
                    \left[r_{\phi^{t-1}_k}(x_i,y_i)-r_{\phi^{t-1}_k}(x_i,b)\right]$
                    \State Query oracle $\mathbb{P}$ on $(x_i,y_i,y_i')$ to obtain $(y_i^+,y_i^-)$
                    \State Update $D_t^O \leftarrow D_t^O \cup \{(x_i,y_i^+,y_i^-)\}$
                    % \State $D_t^O \leftarrow D_t^O \cup \{(x_i,y_i^+,y_i^-)\}$
                \EndIf

            \Statex \hspace{2.7em} {\color{gray}\itshape // Route to ERM branch}
            \ElsIf{$VE(x_i) \le \tau_t^\star$}
                \State $\mathcal{P}_i \leftarrow \{\, (a,b)\in \mathcal{C}_t(x_i),\; a\neq b\;:\; A(x_i,a,b)\le \alpha \,\}$, using Eq.~\eqref{6}

                \If{$\mathcal{P}_i \neq \emptyset$}
                    \Statex \hspace{5.7em} {\color{gray}\itshape // Select most confident pair}
                    \State $(a_i,b_i) \leftarrow \arg\max_{(a,b)\in \mathcal{P}_i}\left|\overline{p}_{ERM^{t-1}}  (a \succ b \mid x_i)-0.5\right|$
                    \State Order $(a_i,b_i)$ as $(\tilde y_i^+, \tilde y_i^-)$ via $\overline{p}_{ERM^{t-1}}$ preference \State Update $D_t^S \leftarrow D_t^S \cup \{(x_i,\tilde y_i^+,\tilde y_i^-)\}$
                    % \State $(\tilde y_i^+, \tilde y_i^-) \leftarrow
                    % \begin{cases}
                    % (a_i,b_i), & \overline{p}_{ERM^{t-1}} (a_i \succ b_i \mid x_i) \geq 0.5\\
                    % (b_i,a_i), & \text{otherwise}
                    % \end{cases}$
                    % \State $D_t^S \leftarrow D_t^S \cup \{(x_i,\tilde y_i^+,\tilde y_i^-)\}$
                \EndIf                
            \EndIf    
            \
        \EndFor
        \State Update ERM parameters $\Phi_t$ using Eq.~\eqref{2} on $\mathcal{D}^O_t$
        \State Update policy parameters $\theta_t$ using Eq.~\eqref{7} on $\mathcal{D}^O_t \cup \mathcal{D}^S_t$         
        
    \EndFor
\end{algorithmic}
\end{algorithm}

%Our vote-entropy routing and variance-based challenger selection are therefore best viewed as practical disagreement-based proxies for epistemic uncertainty: they target model disagreement, but they do not explicitly subtract the conditional-entropy term and hence do not by themselves remove irreducible pairwise ambiguity.%

\subsection{Online Policy Learning}
At each round $t$, for each context $x_i$, we construct a candidate set $\mathcal{C}_t(x_i) = \{y^j\}^N_{j=1}$ by sampling $N = 20$ responses from the latest policy $\pi_{\theta_{t-1}}$. This finite candidate set provides a tractable approximation to searching over the full response space $\mathcal{Y}$, while using the latest policy keeps candidates on-policy and aligned with the current model distribution. 

The preference data produced by the two routing branches serve different roles in the training loop. Oracle-labeled pairs are used to update both the $\pi_{\theta_t}$ and the $\mathcal{R}_{\Phi_t}$, as they provide the ground truth preference signal. ERM-labeled pairs are used only for policy updates and are never fed back into ERM training, since that would cause the ensemble to bootstrap on its own pseudo-labels and gradually corrupt the reward posterior. At the same time, optimizing $\pi_{\theta_t}$ only on oracle data would leave low-disagreement contexts unused, reducing sample efficiency. We therefore train the policy on both sources: oracle comparisons refine uncertain regions, while ERM comparisons provide reliable labels in low-disagreement regions.

To keep AERO agnostic to the direct alignment from preference (DAP) loss, let $F_\theta(x, y^+, y^-)$ denote any DAP objective.\footnote{
Some DAP methods, such as DPO and IPO~\citep{azar2024general}, depend on a
reference policy \(\pi_{\mathrm{ref}}\), while others, such as SimPO~\citep{meng2024simpo}, are reference-free.
}
Given oracle labeled data $\mathcal{D}^O_t$ and ERM labeled data $\mathcal{D}^S_t$ collected at round $t$, we update the policy by minimizing:
\begin{equation}
    \mathcal{L}(\theta \ | \ \mathcal{D}^O_t, \mathcal{D}^S_t, \pi_{ref}) = \mathbb{E}_{(x, y^+, y^-) \sim \mathcal{D}^O_t \cup \mathcal{D}^S_t}  \ [F_{\theta_t}(x, y^+, y^-, \pi_{ref})].
\label{7}
\end{equation}
In our experiments, $F_\theta$ is instantiated as the DPO loss.

\section{Experiments}\label{sec:experiments}
We empirically evaluate AERO against several strong RLHF baselines to assess whether vote-entropy based adaptive routing improves oracle-budget sample efficiency and overall alignment performance. 
Our experiments will answer three important questions: (i) Does AERO achieve higher win rates than competing online RLHF methods at matched oracle budgets? (ii) Does adaptive context routing reduce oracle queries without compromising alignment quality? (iii) What is the contribution of each design choice to AERO's overall performance?

\subsection{Experimental Setup}
\label{sec:setup}
\paragraph{Models and Datasets.} We evaluate AERO on \texttt{Llama-3.2-1B-Instruct}~\citep{grattafiori2024llama}, \texttt{Llama-3-8B-Instruct}, and \texttt{Qwen2.5-1.5B-Instruct} \citep{qwen2024qwen25} to assess whether its gains persist across model scales and families. All experiments use prompts from UltraFeedback Binarized\footnote{The dataset is available at
\url{https://huggingface.co/datasets/HuggingFaceH4/ultrafeedback_binarized}.}, a binarized preprocessing of the original UltraFeedback dataset~\citep{cui2024ultrafeedback}, a widely adopted general instruction-following benchmark for RLHF. We use only the prompts and discard the pre-collected preference labels, since AERO and all baselines generate fresh on-policy responses during training. Holding the prompt distribution fixed allows us to attribute differences in performance primarily to the policy model and exploration dynamics rather than dataset changes. \texttt{Llama-3.2-1B-Instruct} and \texttt{Qwen2.5-1.5B-Instruct} are run over three seeds, whereas \texttt{Llama-3-8B-Instruct} is run on a single seed as a scaling check.

\paragraph{Baselines.} We compare AERO against three online RLHF baselines that span the dominant exploration techniques in the LLM alignment literature. (i) Online DPO~\citep{guo2024direct}, a passive on-policy baseline that samples both responses from the current policy without active exploration; (ii) SEA \citep{liu2024sample}, a fully online active exploration baseline recognized as the closest prior method to AERO and the strongest in terms of sample efficiency. (iii) COPO~\citep{bai2025online}, an optimism-based method that augments DPO with a count-based UCB exploration bonus estimated through a coin-flipping pseudo-count network. Since COPO is not directly designed for our fully on-policy pairwise oracle-budget setting, we use an adapted implementation and provide details in Appendix~\ref{appendix: reproducibility}. Together, the baselines cover posterior-sampling exploration with SEA, optimism-based regularization with COPO, and passive on-policy learning with Online DPO.

\paragraph{Evaluation.} We evaluate checkpoints offline using two complementary metrics. First, we report held-out win rate against the initial instruction-tuned model
under the oracle reward model, plotted against cumulative oracle queries $B$
to measure sample efficiency; and separately report total learner compute and compute costs in Appendix~\ref{app:compute_accounting}. Second, we report length-controlled (LC) and raw win rates on AlpacaEval 2~\citep{dubois2024lengthcontrolled}, evaluated at three checkpoints to assess instruction-following quality under an external judge. Thus, while training is fully online and uses fresh on-policy preference data, evaluation is performed offline on fixed held-out benchmarks.

\subsection{Experimental Results}

\begin{figure}
    \centering
    % First Subfigure
    \begin{subfigure}{0.48\textwidth}
        \centering
        \includegraphics[width=\linewidth]{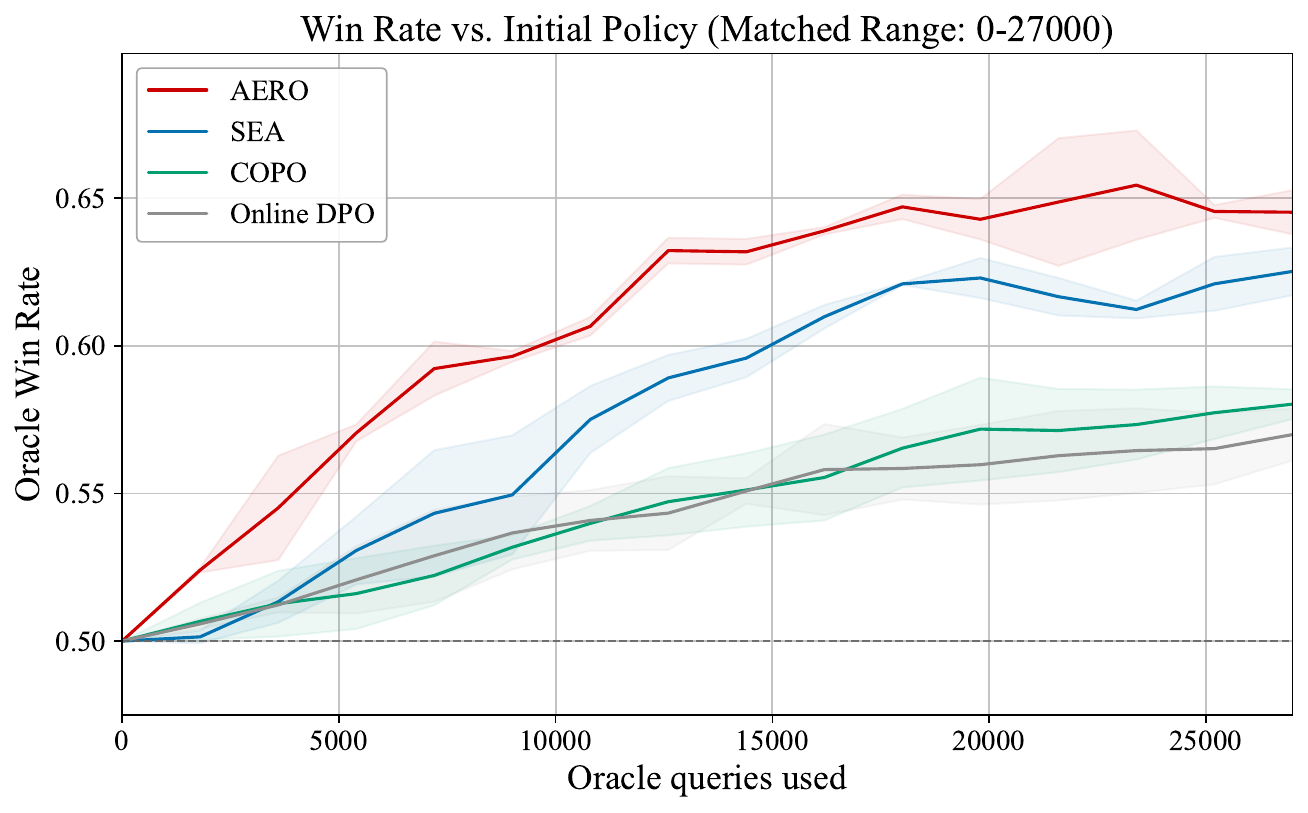}
        \caption{\texttt{Llama-3.2-1B-Instruct}}
        \label{fig:llama_uf_wr_budget}
    \end{subfigure}\hfill
    % Second Subfigure
    \begin{subfigure}{0.48\textwidth}
        \centering
        \includegraphics[width=\linewidth]{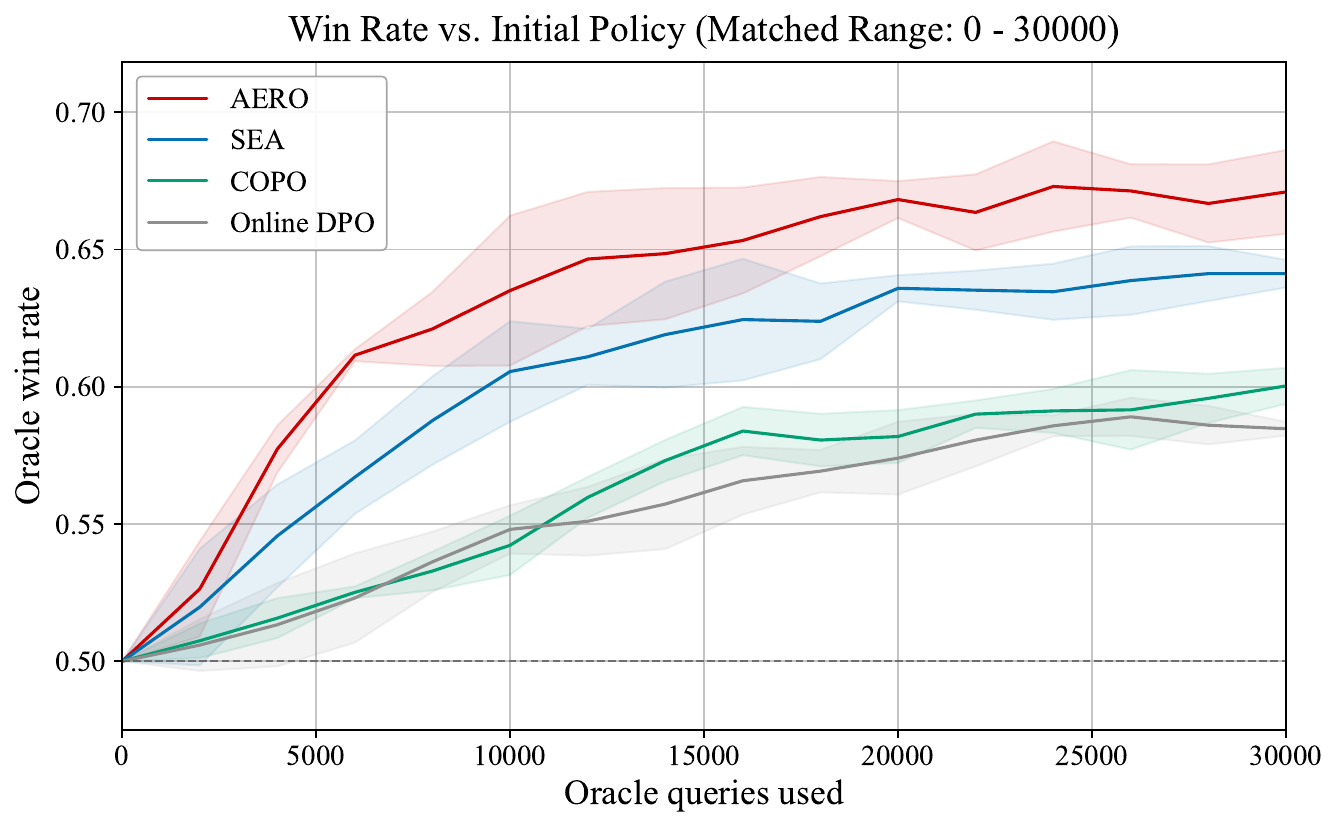}
        \caption{\texttt{Qwen2.5-1.5B-Instruct}}
        \label{fig:qwen15b_uf_wr_budget_30k}
    \end{subfigure}
    
    \caption{Oracle-budget sample efficiency across three seeds. We plot win rate against cumulative oracle queries for (a) \texttt{Llama-3.2-1B-Instruct} and (b) \texttt{Qwen2.5-1.5B-Instruct} on UltraFeedback. Otsu-based adaptive routing automatically determines the routing ratio, which yields different total $B$ across models. The matched budget range is chosen based on AERO’s full $B$ throughout training in each setting.}
    \label{fig:main_results_combined}
\end{figure}

\paragraph{Sample Efficiency Gains.}
\label{sec:sample_efficiency}
\Cref{fig:main_results_combined} shows how win rate against the initial instruction-tuned model scales with cumulative oracle queries under a matched oracle-budget range. Across both \texttt{Llama-3.2-1B-Instruct} and \texttt{Qwen2.5-1.5B-Instruct} on UltraFeedback, AERO achieves the strongest sample-efficiency curve among all methods, demonstrating that the benefit of adaptive ensemble-disagreement routing is not specific to a single model family. On the Llama setup, AERO reaches higher win rates than Online DPO, COPO, and SEA throughout most of training, with the largest gains over the passive Online DPO  and the optimism-based COPO baselines and a smaller but consistent advantage over SEA. The same pattern is observed for \texttt{Qwen2.5-1.5B-Instruct}, where AERO remains above all baselines across the matched 30K oracle-query range, ending with a clear margin over SEA and substantially larger gains over COPO and Online DPO. These results suggest that adaptively routing high-disagreement contexts to the oracle while using ERM-generated labels for low-disagreement contexts improves oracle-budget efficiency without degrading win rate performance and without requiring dataset- or model-specific fixed mixing ratios. Since the main plots use cumulative oracle queries as the x-axis to directly measure sample efficiency, we include iteration-based win-rate curves in Appendix~\ref{Iteration-wise-win-rate-progression} where AERO again shows superior performance. %these are evaluated every 25 training iterations and show the corresponding training-time progression independently of oracle-query consumption where AERO again shows superior performance compared to the baselines.%

\paragraph{Results on AlpacaEval 2.0.}

Table~\ref{tab:alpaca_three_models} reports AlpacaEval 2.0 with LC win rate as the primary AlpacaEval metric, as introduced by~\citet{dubois2024lengthcontrolled} to control for length bias, since LLM-based annotators can exhibit a systematic preference for longer responses. AlpacaEval provides an additional evaluation that is independent of our training oracle, and \citet{dubois2024lengthcontrolled} report that its LC win rate scores achieve a Spearman correlation of 0.98 with LMSYS Chatbot Arena Rankings. AERO achieves its strongest external-judge gains on \texttt{Qwen2.5-1.5B-Instruct}, where it attains the strongest LC and raw win rates at a substantially lower $B$ than the baselines. On \texttt{Llama-3-8B-Instruct}, AERO attains the best final LC, narrowly ahead of SEA at a lower $B$. On \texttt{Llama-3.2-1B-Instruct}, AERO attains the highest mean final LC across three seeds, though it is close enough to SEA and COPO that we do not claim a clear winner. Although COPO achieves a higher raw win rate, its responses are substantially longer on average (1964 vs. 1797 tokens for AERO), suggesting the annotator rewards verbosity due to length bias; thus, LC provides a fairer picture. These results show that AERO's sample-efficiency gains translate to highly competitive instruction-following performance under an external judge.

\begin{table}[t]
\centering
\caption{AlpacaEval 2.0 results across three model settings.
Final LC win rates for 1B and 1.5B class models are mean $\pm$ SE over three seeds for AERO and the two active-exploration baselines; other evaluations are single-run.
$B$ denotes cumulative oracle queries. Iteration indices 1--3 correspond to iterations 100/300/500 for the 1B \& 1.5B class models and 100/200/300 for the 8B model.}
\label{tab:alpaca_three_models}
\vspace{6pt}
\setlength{\tabcolsep}{2.3pt}
\setlength{\aboverulesep}{0pt}
\setlength{\belowrulesep}{0pt}
\renewcommand{\arraystretch}{1.3}
\resizebox{\linewidth}{!}{%
\begin{tabular}{l cccc|cccc|cccc}
\toprule
\multirow{2}{*}{\textbf{Model}}
& \multicolumn{4}{c|}{\textbf{Qwen2.5-1.5B-Instruct}}
& \multicolumn{4}{c|}{\textbf{Llama-3-8B-Instruct}}
& \multicolumn{4}{c}{\textbf{Llama-3.2-1B-Instruct}} \\
\cmidrule(lr){2-5}
\cmidrule(lr){6-9}
\cmidrule(lr){10-13}
& \textbf{LC} & \textbf{WR} & \textbf{Len} & $B$
& \textbf{LC} & \textbf{WR} & \textbf{Len} & $B$
& \textbf{LC} & \textbf{WR} & \textbf{Len} & $B$ \\
\midrule
Base
& 10.14 & 7.38 & 1395 & ---
& 29.71 & 30.15 & 1994 & ---
& 8.06 & 9.23 & 2010 & --- \\
\midrule
DPO 1
& 11.46 & 8.24 & 1389 & 12.8k
& 37.01 & 32.64 & 1766 & 12.8k
& 8.92 & 10.03 & 2019 & 12.8k \\
DPO 2
& 12.34 & 8.62 & 1404 & 38.4k
& 39.48 & 34.66 & 1743 & 25.6k
& 10.16 & 10.99 & 1966 & 38.4k \\
DPO 3
& 12.85 & 8.89 & 1369 & 64.0k
& 40.47 & 34.59 & 1725 & 38.4k
& 8.81 & 9.60 & 1969 & 64.0k \\
\midrule
COPO 1
& 12.10 & 8.89 & 1422 & 12.8k
& 38.64 & 33.36 & 1726 & 12.8k
& 8.61 & 9.72 & 2018 & 12.8k \\
COPO 2
& 12.65 & 8.77 & 1387 & 38.4k
& 39.22 & 33.84 & 1747 & 25.6k
& 10.48 & 11.34 & 1982 & 38.4k \\
COPO 3
& 12.50 {\scriptsize$\pm$} 0.41
& 9.17 {\scriptsize$\pm$} 0.25
& 1411 {\scriptsize$\pm$} 17
& 64.0k
& 38.85 & 33.50 & 1744 & 38.4k
& 11.09 {\scriptsize$\pm$} 0.46
& 11.89 {\scriptsize$\pm$} 0.52
& 1964 {\scriptsize$\pm$} 9
& 64.0k \\
\midrule
SEA 1
& 11.45 & 9.33 & 1613 & 9.3k
& 40.56 & 33.41 & 1645 & 9.3k
& 10.28 & 10.63 & 1906 & 9.3k \\
SEA 2
& 13.34 & 9.57 & 1467 & 27.3k
& 42.07 & 35.74 & 1685 & 18.3k
& 10.75 & 10.40 & 1824 & 27.3k \\
SEA 3
& 14.23 {\scriptsize$\pm$} 0.21
& 10.44 {\scriptsize$\pm$} 0.13
& 1471 {\scriptsize$\pm$} 14
& 45.3k
& 44.36 & 37.81 & 1704 & 27.3k
& 11.54 {\scriptsize$\pm$} 0.50
& 11.03 {\scriptsize$\pm$} 0.57
& 1815 {\scriptsize$\pm$} 7
& 45.3k \\
\midrule
AERO 1
& 13.55 & 11.16 & 1636 & 6.8k
& 40.60 & 33.28 & 1629 & 6.6k
& 9.95 & 10.69 & 1962 & 5.7k \\
AERO 2
& \textbf{16.22} & 12.36 & 1506 & 19.4k
& 40.89 & 34.65 & 1700 & 12.5k
& 10.71 & 10.33 & 1815 & 16.9k \\
AERO 3
& 15.32 {\scriptsize$\pm$} 0.38
& 11.58 {\scriptsize$\pm$} 0.31
& 1480 {\scriptsize$\pm$} 29
& 31.5k
& \textbf{44.39} & 37.36 & 1671 & 18.3k
& \textbf{11.71 {\scriptsize$\pm$} 0.43}
& 11.01 {\scriptsize$\pm$} 0.49
& 1797 {\scriptsize$\pm$} 10
& 27.3k \\
\bottomrule
\end{tabular}%
}
\end{table}

\subsection{Ablation Study}
\label{sec:ablation}

\begin{figure}[t]
    \centering
    % variant1
    \begin{subfigure}{0.48\textwidth}
        \centering
        \includegraphics[width=\linewidth]{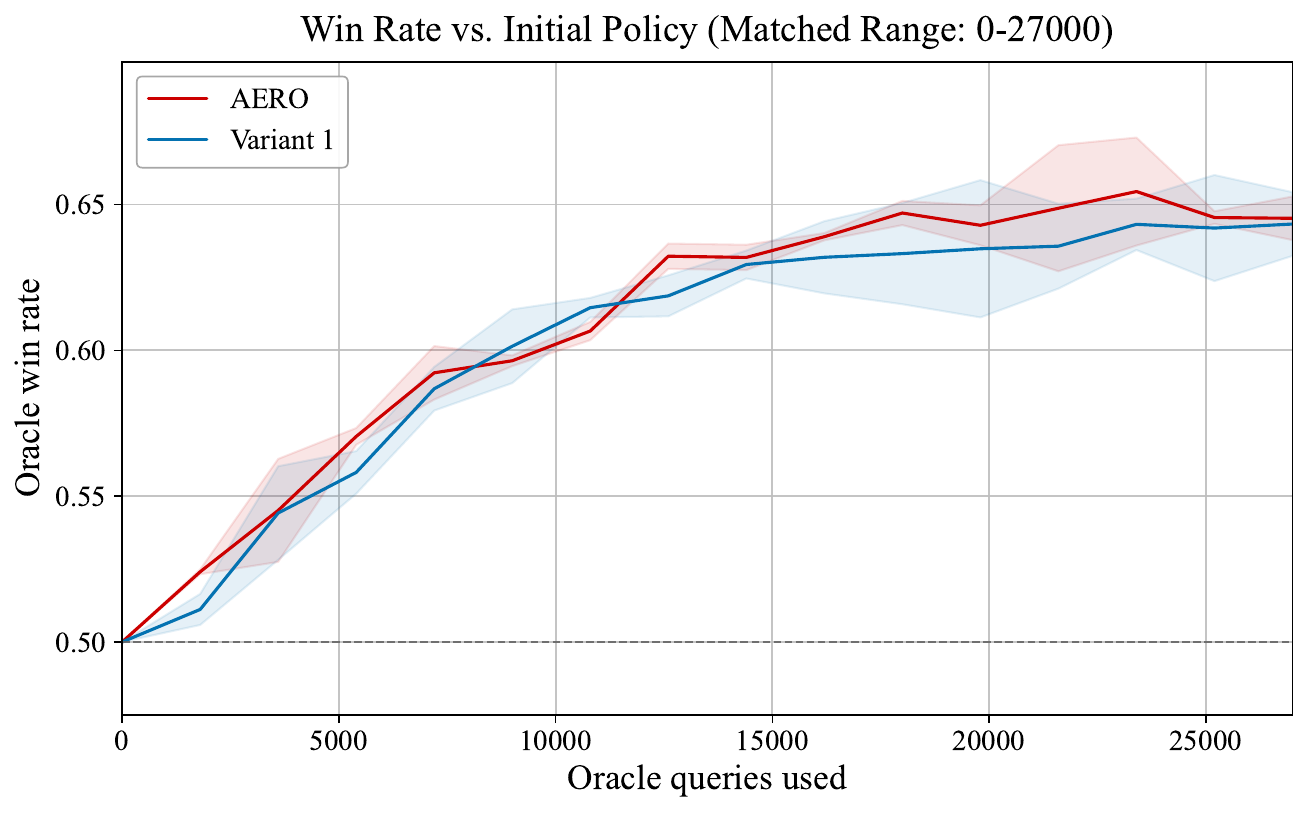}
        \caption{Variant 1: No aleatoric filtering}
        \label{fig:aero_vs_variant1_27k}
    \end{subfigure}%
    \hfill%
    % variant2
    \begin{subfigure}{0.48\textwidth}
        \centering
        \includegraphics[width=\linewidth]{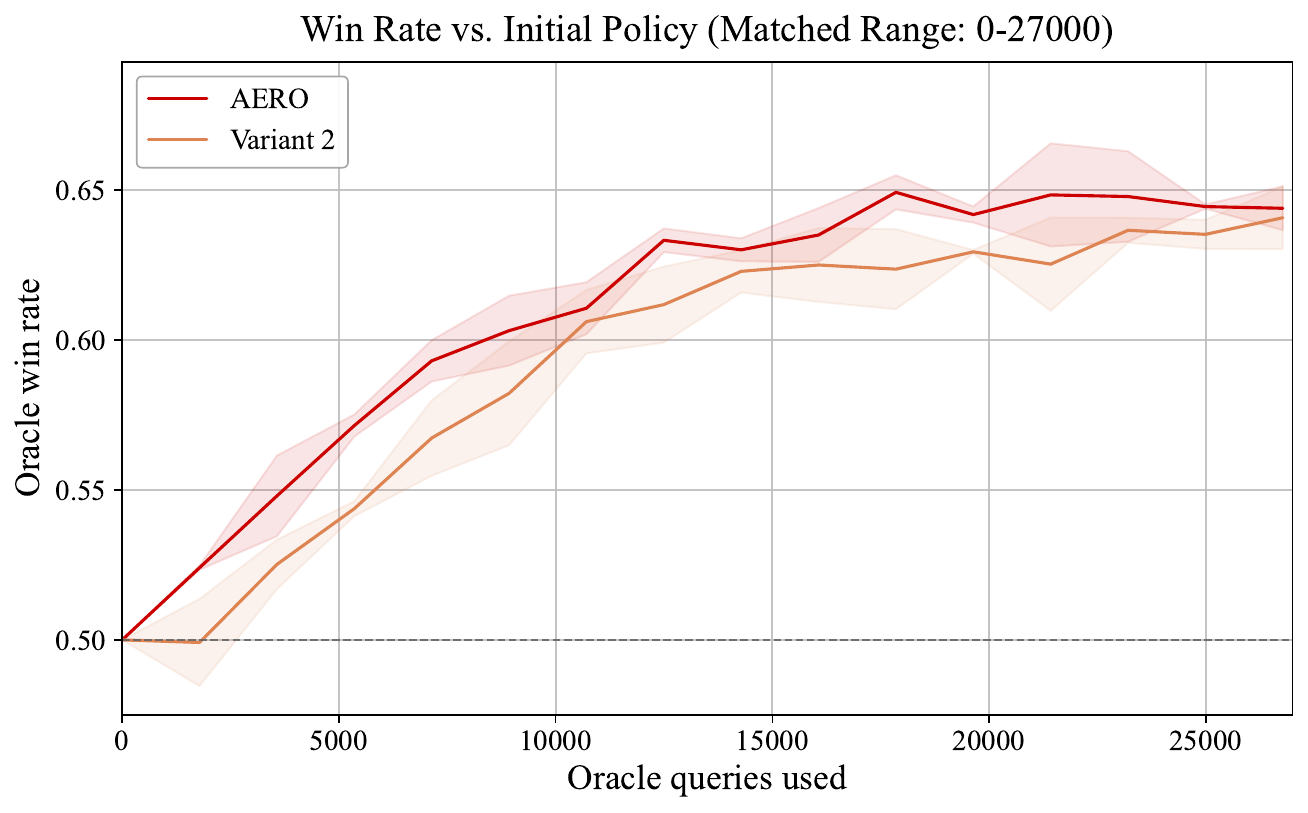}
        \caption{Variant 2: Random Routing}
        \label{fig:aero_vs_variant2_27k}
    \end{subfigure}

    \vspace{0.5em}

    % variant3
    \begin{subfigure}{0.48\textwidth}
        \centering
        \includegraphics[width=\linewidth]{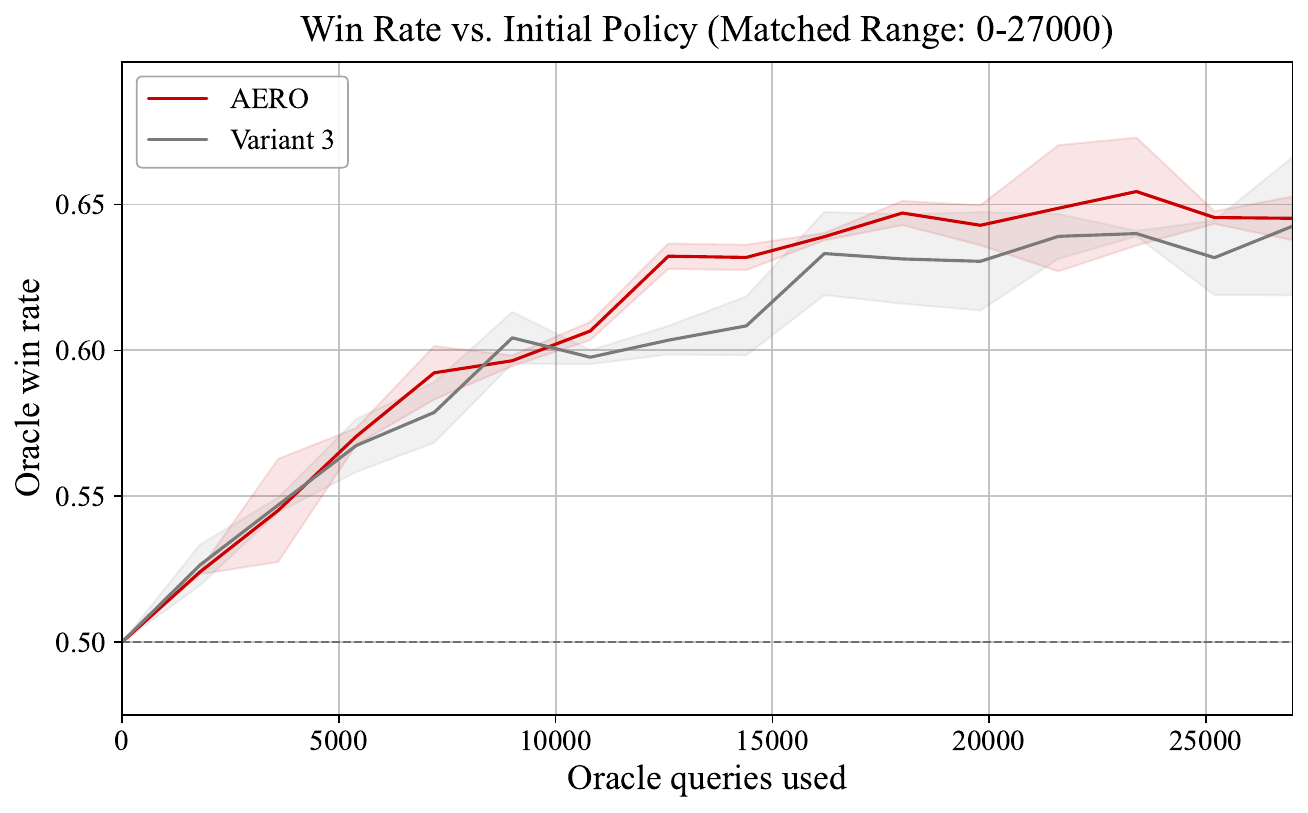}
        \caption{Variant 3: Single-head ERM labeling}
        \label{fig:aero_vs_variant3_27k}
    \end{subfigure}%
    \hfill%
    % variant4
    \begin{subfigure}{0.48\textwidth}
        \centering
        \includegraphics[width=\linewidth]{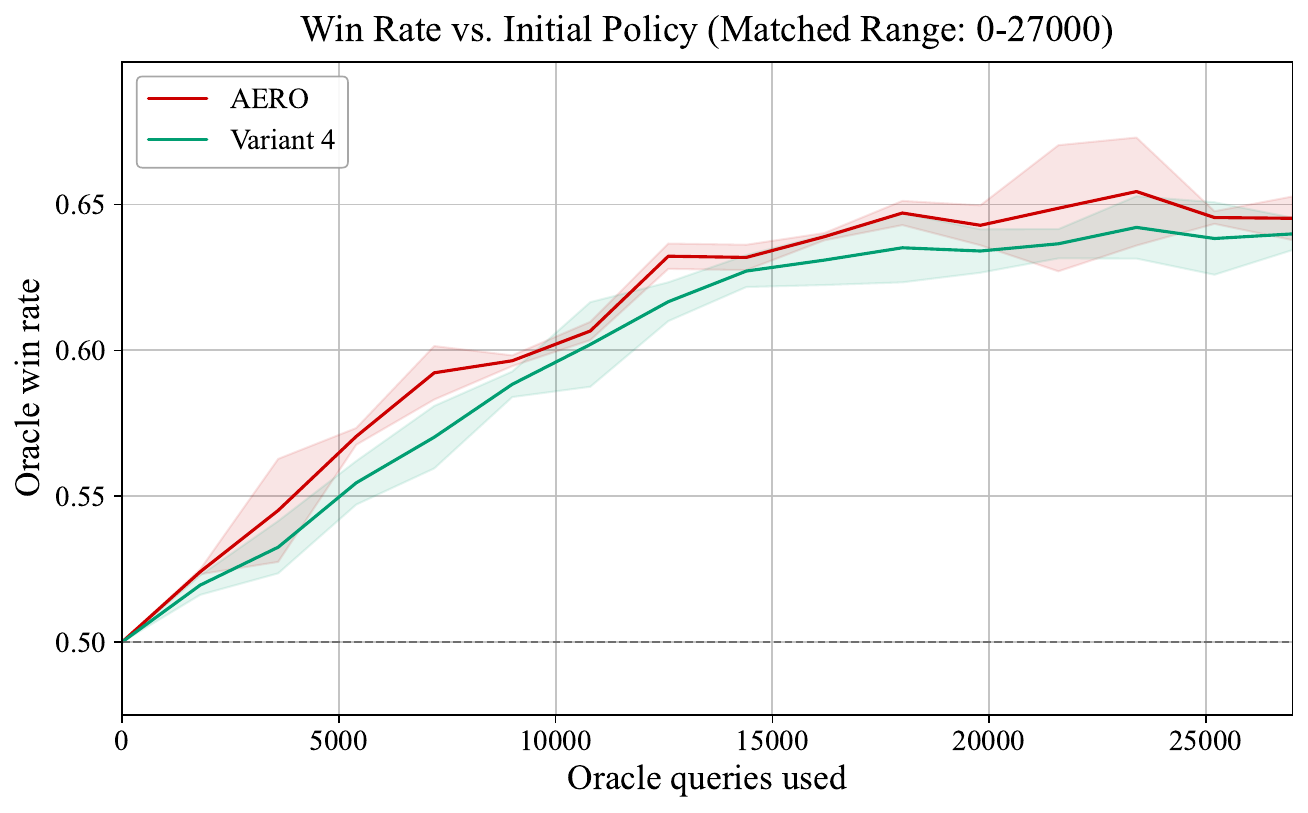}
        \caption{Variant 4: ERM confidence gating}
        \label{fig:aero_vs_variant4_27k}
    \end{subfigure}

    \caption{Ablations on \texttt{Llama-3.2-1B-Instruct} on UltraFeedback. Win rate relative to the initial instruction-tuned model is shown over cumulative oracle-query consumption $B$ for AERO and four single-component ablation variants.}
    \label{fig:ablations_combined}
\end{figure}

We conduct four ablations on the primary setup (\texttt{Llama-3.2-1B-Instruct} on UltraFeedback) to isolate the contribution of each design choice. Each variant modifies a single component while keeping the rest of the algorithm unchanged.
AERO filters out candidate response pairs whose ensemble mean binary entropy exceeds a threshold (Section~\ref{sec:pair selectionn}), removing pairs that are intrinsically ambiguous rather than epistemically uncertain. Variant 1 (\Cref{fig:aero_vs_variant1_27k}) disables this filter, which results in a curve that tracks AERO closely throughout most of training, with a modest separation in the latter half where AERO maintains a small but consistent lead. Variant 2 (\Cref{fig:aero_vs_variant2_27k}) replaces vote-entropy-based routing with random branch assignment at AERO's average realized oracle-routing ratio, while keeping candidate generation, pair selection, aleatoric filtering, ERM labeling, and policy updates unchanged. Across three seeds, AERO achieves a higher mean win rate at every shared oracle budget, and its matched-budget area under the win-rate curve is higher in all three paired runs. This shows that the improvement in oracle budget efficiency does not arise simply from using a lower oracle allocation, but from directing oracle feedback toward high-disagreement contexts. Variant 3 (\Cref{fig:aero_vs_variant3_27k}) replaces ERM's ensemble-mean labeling with SEA's Thompson-sampling-style single-head sampling, motivated by the concern in~\citet{liu2024sample} that averaging across heads collapses epistemic uncertainty. Empirically, Variant 3 tracks AERO closely at low oracle budgets, but from roughly the mid-budget regime onward it remains consistently below AERO for most of training and exhibits a slightly less stable trajectory. This suggests that, within AERO's routing regime, ensemble-mean labeling is a reasonable and stable choice for synthetic supervision, and that the uncertainty-collapse concern is less consequential once synthetic candidates have already been filtered to high-consensus contexts. Variant 4 (\Cref{fig:aero_vs_variant4_27k}) adds a confidence-gated oracle fallback on top of adaptive routing, redirecting contexts whose most confident ERM-labeled pair falls below the gate to oracle supervision. We include it as a robustness check to prevent the ERM from labeling low-confidence pairs on the assumption that oracle supervision might be more useful. Empirically, Variant 4 remains slightly but consistently below AERO throughout training while consuming more oracle queries ($B \approx 38$K), indicating that the added confidence gate does not improve over vote-entropy-based routing alone and can reduce oracle efficiency. Together, these ablations indicate that some components contribute incrementally, while majority of sample-efficiency gains come from adaptive vote-entropy based routing.

\section{Related Work}

\textbf{Online Preference Alignment} addresses the severe distribution shift and limited data coverage in standard offline alignment methods. Direct alignment from preference (DAP) methods~\citep{rafailov2023direct, meng2024simpo, pmlr-v235-ethayarajh24a, azar2024general, xu2024contrastive} have become widely adopted alternatives to PPO-based RLHF, but these offline methods rely on static datasets where responses were generated by a reference model prior to RLHF training and are bottlenecked by the coverage of that dataset, leading to distributional shift and overoptimization on out-of-distribution responses~\citep{xiong2024iterative}.
Many online methods~\citep{xu2023some, xiong2024iterative, dong2024rlhf} periodically refresh the preference dataset across a small number of iterations by sampling new responses from the latest policy checkpoint and relabeling them with a preference oracle.~\citet{tajwar2024preference} and~\citet{tang2024understanding} argue that online alignment's empirical advantage is driven by on-policy preference data, motivating fully online variants such as OAIF~\citep{guo2024direct} where a fresh on-policy preference pair is collected at every gradient step. %However, while these frameworks resolve off-policy drift, their reliance on uniform random context sampling leaves a gap in sample efficiency that targeted active exploration can close.%

\textbf{Active Preference Exploration} is a growing line of work that  incorporates active exploration in the LLM alignment process by strategically selecting prompts or response pairs that maximize the information gained per oracle query. A common paradigm is pool-based active learning, where an acquisition function scores a pre-collected pool of prompts or response pairs and selects only the highest-scoring subset for oracle labeling. APL~\citep{muldrew2024active} follows this by sampling a large batch of contexts, generating completions from the current model, scoring action pairs conditioned on a context using language-model entropy and DPO implicit preference certainty, querying the oracle on the top-ranked subset. APO~\citep{das2025active} similarly selects a batch of high-uncertainty context-generation pairs from a finite set. BAL-PM~\citep{melo2024deep} focuses specifically on Active Preference Modeling, using a Bayesian stochastic acquisition policy that selects examples with high preference-model epistemic uncertainty while encouraging prompt diversity through an entropy term. A complementary line of work integrates exploration into preference-data acquisition for online alignment. For instance,~\citet{mehta2025sample} propose an active DPO-style acquisition strategy using uncertainty-based context selection and online policy updates, while~\citet{dwaracherla2024efficient} use an epistemic reward model with double Thompson sampling to select response pairs, focusing on reward-model learning and best-of-N response selection rather than online policy optimization. SEA~\citep{liu2024sample} is the closest prior work combining fully online policy updates with an ensemble ERM for posterior sampling, TS-style response selection, and Dyna-style mixed preference learning. AERO therefore has a computational profile similar to SEA.

A separate line of work incorporates exploration directly into the policy optimization objective rather than using an acquisition strategy. SELM~\citep{zhang2024self} and XPO \citep{xie2025xpo} encourage exploration by augmenting the DPO objective with an optimistic regularization term that biases the implicit reward model towards potentially high-reward, out-of-distribution responses. VPO~\citep{cenvalue} adopts a similar perspective, incorporating the maximum of the KL-regularized value function into the reward objective for unified optimism across online and offline RLHF. COPO~\citep{bai2025online} follows this paradigm with a count-based UCB exploration bonus term added to the DPO loss, estimated through a coin-flipping pseudo-count network~\citep{lobel2023flipping}.
 
\textbf{Query-by-Committee} is a classical active learning framework in which examples are selected for annotation based on disagreement among a committee of models. Vote entropy is a simple and widely used measure of committee disagreement in active learning, with prior applications to probabilistic classifiers and NLP annotation tasks such as part-of-speech tagging, sequence labeling, dependency parsing, and coreference resolution~\citep{seung1992query, freund1997selective, argamon1999committee, engelson1996minimizing, dagan1995selective, settles2008analysis, majidi2013active, li2020active}. AERO builds on this line of work, but uses vote entropy differently: rather than selecting static unlabeled examples from a pool, it scores contexts using ERM ensemble disagreement and uses this signal to decide whether each context should receive oracle feedback or ERM-generated preference labels.

\section{Closing Remarks}\label{sec:conclusion}
We presented AERO, a fully online RLHF algorithm that adaptively allocates oracle supervision across contexts based on ensemble disagreement. By using vote entropy over a deep ensemble ERM as a query-by-committee routing signal, combined with Otsu-based adaptive thresholding, AERO concentrates expensive oracle queries on high-disagreement contexts while still training the policy on low-disagreement contexts through ERM synthetic labels, so that the long tail of confident contexts is not wasted. Empirically, AERO achieves highly competitive
alignment performance compared to Online DPO and strong active-exploration
baselines while reducing the required oracle-query budget across two model
families spanning the 1B to 8B range. Our ablations show that aleatoric filtering and ensemble-mean synthetic
labeling provide small but consistent gains, while the matched-budget random-routing
control confirms that the primary sample-efficiency gains arise from
directing oracle feedback toward high-disagreement contexts rather than
simply reducing the overall oracle allocation.

\paragraph{Limitations \& Future Work.} We use a deterministic reward model as the preference oracle following prior work. This isolates algorithmic differences from oracle noise but does not capture the stochasticity of real human feedback. Therefore, evaluating AERO using noisy or real human feedback is an important direction for future work. Our experiments also focus on the common research setting in this line of work where annotators are paid to provide preference labels for training~\citep{christiano2017deep, ouyang2022training} and policies are evaluated afterwards. We do not study the deployment setting in which a live LLM application is continually aligned with its users' feedback, as in commercial systems that ask users to compare two responses. Finally, although AERO is agnostic to the choice of direct alignment loss (Eq.~\ref{7}), our experiments instantiate it only with DPO. Hence, evaluating AERO with other DAP objectives such as IPO~\citep{azar2024general} and SimPO~\citep{meng2024simpo} remains important future work.

\begin{ack}
We thank the RMIT AWS Cloud Supercomputing Hub (RACE) for partially providing the compute required to run the experiments in this paper. The authors declare no competing interests.
\end{ack}

\bibliographystyle{plainnat}
\bibliography{references}

@article{liu2024sample,
  title={Sample-efficient alignment for llms},
  author={Liu, Zichen and Chen, Changyu and Du, Chao and Lee, Wee Sun and Lin, Min},
  journal={arXiv preprint arXiv:2411.01493},
  year={2024}
}

@article{christiano2017deep,
  title={Deep reinforcement learning from human preferences},
  author={Christiano, Paul F and Leike, Jan and Brown, Tom and Martic, Miljan and Legg, Shane and Amodei, Dario},
  journal={Advances in neural information processing systems},
  volume={30},
  year={2017}
}

@article{bai2022training,
  title={Training a helpful and harmless assistant with reinforcement learning from human feedback},
  author={Bai, Yuntao and Jones, Andy and Ndousse, Kamal and Askell, Amanda and Chen, Anna and DasSarma, Nova and Drain, Dawn and Fort, Stanislav and Ganguli, Deep and Henighan, Tom and others},
  journal={arXiv preprint arXiv:2204.05862},
  year={2022}
}

@article{ouyang2022training,
  title={Training language models to follow instructions with human feedback},
  author={Ouyang, Long and Wu, Jeffrey and Jiang, Xu and Almeida, Diogo and Wainwright, Carroll and Mishkin, Pamela and Zhang, Chong and Agarwal, Sandhini and Slama, Katarina and Ray, Alex and others},
  journal={Advances in neural information processing systems},
  volume={35},
  pages={27730--27744},
  year={2022}
}

@article{schulman2017proximal,
  title={Proximal policy optimization algorithms},
  author={Schulman, John and Wolski, Filip and Dhariwal, Prafulla and Radford, Alec and Klimov, Oleg},
  journal={arXiv preprint arXiv:1707.06347},
  year={2017}
}

@article{rafailov2023direct,
  title={Direct preference optimization: Your language model is secretly a reward model},
  author={Rafailov, Rafael and Sharma, Archit and Mitchell, Eric and Manning, Christopher D and Ermon, Stefano and Finn, Chelsea},
  journal={Advances in neural information processing systems},
  volume={36},
  pages={53728--53741},
  year={2023}
}

@inproceedings{luo2025survey,
  title={A survey on efficient large language model training: From data-centric perspectives},
  author={Luo, Junyu and Wu, Bohan and Luo, Xiao and Xiao, Zhiping and Jin, Yiqiao and Tu, Rong-Cheng and Yin, Nan and Wang, Yifan and Yuan, Jingyang and Ju, Wei and others},
  booktitle={Proceedings of the 63rd Annual Meeting of the Association for Computational Linguistics (Volume 1: Long Papers)},
  pages={30904--30920},
  year={2025}
}

@inproceedings{das2025active,
  title={Active preference optimization for sample efficient rlhf},
  author={Das, Nirjhar and Chakraborty, Souradip and Pacchiano, Aldo and Chowdhury, Sayak Ray},
  booktitle={Joint European Conference on Machine Learning and Knowledge Discovery in Databases},
  pages={96--112},
  year={2025},
  organization={Springer}
}

@inproceedings{muldrew2024active,
  title={Active Preference Learning for Large Language Models},
  author={Muldrew, William and Hayes, Peter and Zhang, Mingtian and Barber, David},
  booktitle={International Conference on Machine Learning},
  pages={36577--36590},
  year={2024},
  organization={PMLR}
}

@article{zhang2024self,
  title={Self-exploring language models: Active preference elicitation for online alignment},
  author={Zhang, Shenao and Yu, Donghan and Sharma, Hiteshi and Zhong, Han and Liu, Zhihan and Yang, Ziyi and Wang, Shuohang and Hassan, Hany and Wang, Zhaoran},
  journal={arXiv preprint arXiv:2405.19332},
  year={2024}
}

@inproceedings{dwaracherla2024efficient,
  title={Efficient Exploration for LLMs},
  author={Dwaracherla, Vikranth and Asghari, Seyed Mohammad and Hao, Botao and Van Roy, Benjamin},
  booktitle={International Conference on Machine Learning},
  pages={12215--12227},
  year={2024},
  organization={PMLR}
}

@inproceedings{
bai2025online,
title={Online Preference Alignment for Language Models via Count-based Exploration},
author={Chenjia Bai and Yang Zhang and Shuang Qiu and Qiaosheng Zhang and Kang Xu and Xuelong Li},
booktitle={The Thirteenth International Conference on Learning Representations},
year={2025},
}

@article{argamon1999committee,
  title={Committee-based sample selection for probabilistic classifiers},
  author={Argamon-Engelson, Shlomo and Dagan, Ido},
  journal={Journal of Artificial Intelligence Research},
  volume={11},
  pages={335--360},
  year={1999}
}

@article{freund1997selective,
  title={Selective sampling using the query by committee algorithm},
  author={Freund, Yoav and Seung, H Sebastian and Shamir, Eli and Tishby, Naftali},
  journal={Machine learning},
  volume={28},
  number={2},
  pages={133--168},
  year={1997},
  publisher={Springer}
}

@incollection{sutton1990integrated,
  title={Integrated architectures for learning, planning, and reacting based on approximating dynamic programming},
  author={Sutton, Richard S},
  booktitle={Machine learning proceedings 1990},
  pages={216--224},
  year={1990},
  publisher={Elsevier}
}

@article{janner2019trust,
  title={When to trust your model: Model-based policy optimization},
  author={Janner, Michael and Fu, Justin and Zhang, Marvin and Levine, Sergey},
  journal={Advances in neural information processing systems},
  volume={32},
  year={2019}
}

@article{asghari2026efficient,
  title={Efficient Exploration at Scale},
  author={Asghari, Seyed Mohammad and Chute, Chris and Dwaracherla, Vikranth and Lu, Xiuyuan and Jafarnia, Mehdi and Minden, Victor and Wen, Zheng and Van Roy, Benjamin},
  journal={arXiv preprint arXiv:2603.17378},
  year={2026}
}

@article{saha2021optimal,
  title={Optimal algorithms for stochastic contextual preference bandits},
  author={Saha, Aadirupa},
  journal={Advances in Neural Information Processing Systems},
  volume={34},
  pages={30050--30062},
  year={2021}
}

@inproceedings{dudik2015contextual,
  title={Contextual dueling bandits},
  author={Dud{\'\i}k, Miroslav and Hofmann, Katja and Schapire, Robert E and Slivkins, Aleksandrs and Zoghi, Masrour},
  booktitle={Conference on Learning Theory},
  pages={563--587},
  year={2015},
  organization={PMLR}
}

@article{bradley1952rank,
  title={Rank analysis of incomplete block designs: I. the method of paired comparisons},
  author={Bradley, Ralph Allan and Terry, Milton E},
  journal={Biometrika},
  volume={39},
  number={3/4},
  pages={324--345},
  year={1952},
  publisher={JSTOR}
}

@article{lakshminarayanan2017simple,
  title={Simple and scalable predictive uncertainty estimation using deep ensembles},
  author={Lakshminarayanan, Balaji and Pritzel, Alexander and Blundell, Charles},
  journal={Advances in neural information processing systems},
  volume={30},
  year={2017}
}

@article{otsu1979threshold,
  author  = {Nobuyuki Otsu},
  title   = {A Threshold Selection Method from Gray-Level Histograms},
  journal = {IEEE Transactions on Systems, Man, and Cybernetics},
  volume  = {9},
  number  = {1},
  pages   = {62--66},
  year    = {1979},
  doi     = {10.1109/TSMC.1979.4310076}
}

@inproceedings{depeweg2018decomposition,
  title={Decomposition of uncertainty in Bayesian deep learning for efficient and risk-sensitive learning},
  author={Depeweg, Stefan and Hernandez-Lobato, Jose-Miguel and Doshi-Velez, Finale and Udluft, Steffen},
  booktitle={International conference on machine learning},
  pages={1184--1193},
  year={2018},
  organization={PMLR}
}

@article{thompson1933likelihood,
  title={On the likelihood that one unknown probability exceeds another in view of the evidence of two samples},
  author={Thompson, William R},
  journal={Biometrika},
  volume={25},
  number={3/4},
  pages={285--294},
  year={1933},
  publisher={JSTOR}
}

@article{melo2024deep,
  title={Deep bayesian active learning for preference modeling in large language models},
  author={Melo, Luckeciano C and Tigas, Panagiotis and Abate, Alessandro and Gal, Yarin},
  journal={Advances in Neural Information Processing Systems},
  volume={37},
  pages={118052--118085},
  year={2024}
}

@article{houlsby2011bayesian,
  title={Bayesian active learning for classification and preference learning},
  author={Houlsby, Neil and Husz{\'a}r, Ferenc and Ghahramani, Zoubin and Lengyel, M{\'a}t{\'e}},
  journal={arXiv preprint arXiv:1112.5745},
  year={2011}
}

@inproceedings{seung1992query,
  title={Query by committee},
  author={Seung, H Sebastian and Opper, Manfred and Sompolinsky, Haim},
  booktitle={Proceedings of the fifth annual workshop on Computational learning theory},
  pages={287--294},
  year={1992}
}

@techreport{settles2009active,
Author = {Burr Settles},
Institution = {University of Wisconsin--Madison},
Number = {1648},
Title = {Active Learning Literature Survey},
Type = {Computer Sciences Technical Report},
Year = {2009},
}

@article{kirsch2019batchbald,
  title={Batchbald: Efficient and diverse batch acquisition for deep bayesian active learning},
  author={Kirsch, Andreas and Van Amersfoort, Joost and Gal, Yarin},
  journal={Advances in neural information processing systems},
  volume={32},
  year={2019}
}

@article{der2009aleatory,
  title={Aleatory or epistemic? Does it matter?},
  author={Der Kiureghian, Armen and Ditlevsen, Ove},
  journal={Structural safety},
  volume={31},
  number={2},
  pages={105--112},
  year={2009},
  publisher={Elsevier}
}

@article{kendall2017uncertainties,
  title={What uncertainties do we need in bayesian deep learning for computer vision?},
  author={Kendall, Alex and Gal, Yarin},
  journal={Advances in neural information processing systems},
  volume={30},
  year={2017}
}

@article{cohn1996active,
  title={Active learning with statistical models},
  author={Cohn, David A and Ghahramani, Zoubin and Jordan, Michael I},
  journal={Journal of artificial intelligence research},
  volume={4},
  pages={129--145},
  year={1996}
}

@inproceedings{azar2024general,
  title={A general theoretical paradigm to understand learning from human preferences},
  author={Azar, Mohammad Gheshlaghi and Guo, Zhaohan Daniel and Piot, Bilal and Munos, Remi and Rowland, Mark and Valko, Michal and Calandriello, Daniele},
  booktitle={International Conference on Artificial Intelligence and Statistics},
  pages={4447--4455},
  year={2024},
  organization={PMLR}
}

@article{meng2024simpo,
  title={Simpo: Simple preference optimization with a reference-free reward},
  author={Meng, Yu and Xia, Mengzhou and Chen, Danqi},
  journal={Advances in Neural Information Processing Systems},
  volume={37},
  pages={124198--124235},
  year={2024}
}

@inproceedings{yu2020multi,
  title={Multi-task curriculum framework for open-set semi-supervised learning},
  author={Yu, Qing and Ikami, Daiki and Irie, Go and Aizawa, Kiyoharu},
  booktitle={European conference on computer vision},
  pages={438--454},
  year={2020},
  organization={Springer}
}

@inproceedings{ma2023rethinking,
  title={Rethinking safe semi-supervised learning: Transferring the open-set problem to a close-set one},
  author={Ma, Qiankun and Gao, Jiyao and Zhan, Bo and Guo, Yunpeng and Zhou, Jiliu and Wang, Yan},
  booktitle={Proceedings of the IEEE/CVF International Conference on Computer Vision},
  pages={16370--16379},
  year={2023}
}

@inproceedings{engelson1996minimizing,
  title={Minimizing manual annotation cost in supervised training from corpora},
  author={Engelson, Sean P and Dagan, Ido},
  booktitle={34th Annual Meeting of the Association for Computational Linguistics},
  pages={319--326},
  year={1996}
}

@inproceedings{dagan1995selective,
  title={Selective sampling in natural language learning},
  author={Dagan, Ido and Engelson, Sean P},
  booktitle={Proceedings of the IJCAI Workshop on New Approaches to Learning for Natural Language Processing},
  pages={41--48},
  year={1995}
}

@inproceedings{settles2008analysis,
  title={An analysis of active learning strategies for sequence labeling tasks},
  author={Settles, Burr and Craven, Mark},
  booktitle={proceedings of the 2008 conference on empirical methods in natural language processing},
  pages={1070--1079},
  year={2008}
}

@inproceedings{majidi2013active,
  title={Active learning for dependency parsing by a committee of parsers},
  author={Majidi, Saeed and Crane, Gregory},
  booktitle={Proceedings of the 13th International Conference on Parsing Technologies (IWPT 2013)},
  pages={98--105},
  year={2013}
}

@inproceedings{li2020active,
  title={Active learning for coreference resolution using discrete annotation},
  author={Li, Belinda Z and Stanovsky, Gabriel and Zettlemoyer, Luke},
  booktitle={Proceedings of the 58th Annual Meeting of the Association for Computational Linguistics},
  pages={8320--8331},
  year={2020}
}

@InProceedings{pmlr-v235-ethayarajh24a,
  title = 	 {Model Alignment as Prospect Theoretic Optimization},
  author =       {Ethayarajh, Kawin and Xu, Winnie and Muennighoff, Niklas and Jurafsky, Dan and Kiela, Douwe},
  booktitle = 	 {Proceedings of the 41st International Conference on Machine Learning},
  pages = 	 {12634--12651},
  year = 	 {2024},
  editor = 	 {Salakhutdinov, Ruslan and Kolter, Zico and Heller, Katherine and Weller, Adrian and Oliver, Nuria and Scarlett, Jonathan and Berkenkamp, Felix},
  volume = 	 {235},
  series = 	 {Proceedings of Machine Learning Research},
  month = 	 {21--27 Jul},
  publisher =    {PMLR},
}

@inproceedings{
    xu2024contrastive,
    title={Contrastive Preference Optimization: Pushing the Boundaries of {LLM} Performance in Machine Translation},
    author={Haoran Xu and Amr Sharaf and Yunmo Chen and Weiting Tan and Lingfeng Shen and Benjamin Van Durme and Kenton Murray and Young Jin Kim},
    booktitle={Forty-first International Conference on Machine Learning},
    year={2024}
}

@inproceedings{
    xiong2024iterative,
    title={Iterative Preference Learning from Human Feedback: Bridging Theory and Practice for {RLHF} under {KL}-constraint},
    author={Wei Xiong and Hanze Dong and Chenlu Ye and Ziqi Wang and Han Zhong and Heng Ji and Nan Jiang and Tong Zhang},
    booktitle={Forty-first International Conference on Machine Learning},
    year={2024}
}

@article{xu2023some,
  title={Some things are more cringe than others: Iterative preference optimization with the pairwise cringe loss},
  author={Xu, Jing and Lee, Andrew and Sukhbaatar, Sainbayar and Weston, Jason},
  journal={arXiv preprint arXiv:2312.16682},
  year={2023}
}

@article{
    dong2024rlhf,
    title={{RLHF} Workflow: From Reward Modeling to Online {RLHF}},
    author={Hanze Dong and Wei Xiong and Bo Pang and Haoxiang Wang and Han Zhao and Yingbo Zhou and Nan Jiang and Doyen Sahoo and Caiming Xiong and Tong Zhang},
    journal={Transactions on Machine Learning Research},
    issn={2835-8856},
    year={2024},
    note={}
}

@inproceedings{tajwar2024preference,
  title={Preference Fine-Tuning of LLMs Should Leverage Suboptimal, On-Policy Data},
  author={Tajwar, Fahim and Singh, Anikait and Sharma, Archit and Rafailov, Rafael and Schneider, Jeff and Xie, Tengyang and Ermon, Stefano and Finn, Chelsea and Kumar, Aviral},
  booktitle={International Conference on Machine Learning},
  pages={47441--47474},
  year={2024},
  organization={PMLR}
}

@article{tang2024understanding,
  title={Understanding the performance gap between online and offline alignment algorithms},
  author={Tang, Yunhao and Guo, Daniel Zhaohan and Zheng, Zeyu and Calandriello, Daniele and Cao, Yuan and Tarassov, Eugene and Munos, R{\'e}mi and Pires, Bernardo {\'A}vila and Valko, Michal and Cheng, Yong and others},
  journal={arXiv preprint arXiv:2405.08448},
  year={2024}
}

@article{guo2024direct,
  title={Direct language model alignment from online ai feedback},
  author={Guo, Shangmin and Zhang, Biao and Liu, Tianlin and Liu, Tianqi and Khalman, Misha and Llinares, Felipe and Rame, Alexandre and Mesnard, Thomas and Zhao, Yao and Piot, Bilal and others},
  journal={arXiv preprint arXiv:2402.04792},
  year={2024}
}

@inproceedings{mehta2025sample,
  title={Sample Efficient Preference Alignment in {LLM}s via Active Exploration},
  author={Mehta, Viraj and Belakaria, Syrine and Das, Vikramjeet and Neopane, Ojash and Dai, Yijia and Bogunovic, Ilija and Engelhardt, Barbara and Ermon, Stefano and Schneider, Jeff and Neiswanger, Willie},
  booktitle={Second Conference on Language Modeling},
  year={2025}
}

@inproceedings{xie2025xpo,
  title = {Exploratory Preference Optimization: Harnessing Implicit {Q*}-Approximation for Sample-Efficient {RLHF}},
  author = {Xie, Tengyang and Foster, Dylan J. and Krishnamurthy, Akshay and Rosset, Corby and Awadallah, Ahmed and Rakhlin, Alexander},
  booktitle = {International Conference on Learning Representations},
  year = {2025}
}

@inproceedings{cenvalue,
  title = {Value-Incentivized Preference Optimization: A Unified Approach to Online and Offline {RLHF}},
  author = {Cen, Shicong and Mei, Jincheng and Goshvadi, Katayoon and Dai, Hanjun and Yang, Tong and Yang, Sherry and Schuurmans, Dale and Chi, Yuejie and Dai, Bo},
  booktitle = {International Conference on Learning Representations},
  year  = {2025},
}

@inproceedings{lobel2023flipping,
  title={Flipping coins to estimate pseudocounts for exploration in reinforcement learning},
  author={Lobel, Sam and Bagaria, Akhil and Konidaris, George},
  booktitle={International Conference on Machine Learning},
  pages={22594--22613},
  year={2023},
  organization={PMLR}
}

@article{grattafiori2024llama,
  title={The llama 3 herd of models},
  author={Grattafiori, Aaron and Dubey, Abhimanyu and Jauhri, Abhinav and Pandey, Abhinav and Kadian, Abhishek and Al-Dahle, Ahmad and Letman, Aiesha and Mathur, Akhil and Schelten, Alan and Vaughan, Alex and others},
  journal={arXiv preprint arXiv:2407.21783},
  year={2024}
}

@inproceedings{jiang2023llm,
  title={Llm-blender: Ensembling large language models with pairwise ranking and generative fusion},
  author={Jiang, Dongfu and Ren, Xiang and Lin, Bill Yuchen},
  booktitle={Proceedings of the 61st Annual Meeting of the Association for Computational Linguistics (Volume 1: Long Papers)},
  pages={14165--14178},
  year={2023}
}

@inproceedings{cui2024ultrafeedback,
  title={ULTRAFEEDBACK: Boosting Language Models with Scaled AI Feedback},
  author={Cui, Ganqu and Yuan, Lifan and Ding, Ning and Yao, Guanming and He, Bingxiang and Zhu, Wei and Ni, Yuan and Xie, Guotong and Xie, Ruobing and Lin, Yankai and others},
  booktitle={International Conference on Machine Learning},
  pages={9722--9744},
  year={2024},
  organization={PMLR}
}

@article{qwen2024qwen25,
  title   = {{Qwen2.5} Technical Report},
  author={Qwen An Yang and Baosong Yang and Beichen Zhang and Binyuan Hui and Bo Zheng and Bowen Yu and Chengyuan Li and Dayiheng Liu and Fei Huang and Guanting Dong and Haoran Wei and Huan Lin and Jian Yang and Jianhong Tu and Jianwei Zhang and Jianxin Yang and Jiaxin Yang and Jingren Zhou and Junyang Lin and Kai Dang and Keming Lu and Keqin Bao and Kexin Yang and Le Yu and Mei Li and Mingfeng Xue and Pei Zhang and Qin Zhu and Rui Men and Runji Lin and Tianhao Li and Tingyu Xia and Xingzhang Ren and Xuancheng Ren and Yang Fan and Yang Su and Yi-Chao Zhang and Yunyang Wan and Yuqi Liu and Zeyu Cui and Zhenru Zhang and Zihan Qiu and Shanghaoran Quan and Zekun Wang},
  journal = {arXiv preprint arXiv:2412.15115},
  year    = {2024}
}

@article{liu2024skywork,
  title={Skywork-reward: Bag of tricks for reward modeling in llms},
  author={Liu, Chris Yuhao and Zeng, Liang and Liu, Jiacai and Yan, Rui and He, Jujie and Wang, Chaojie and Yan, Shuicheng and Liu, Yang and Zhou, Yahui},
  journal={arXiv preprint arXiv:2410.18451},
  year={2024}
}

@inproceedings{
dubois2024lengthcontrolled,
title={Length-Controlled AlpacaEval: A Simple Debiasing of Automatic Evaluators},
author={Yann Dubois and Percy Liang and Tatsunori Hashimoto},
booktitle={First Conference on Language Modeling},
year={2024},
}

@inproceedings{
    hu2022lora,
    title={Lo{RA}: Low-Rank Adaptation of Large Language Models},
    author={Edward J Hu and yelong shen and Phillip Wallis and Zeyuan Allen-Zhu and Yuanzhi Li and Shean Wang and Lu Wang and Weizhu Chen},
    booktitle={International Conference on Learning Representations},
    year={2022},
}

@inproceedings{volske2017tldr,
  title = {{TL;DR}: Mining Reddit to Learn Automatic Summarization},
  author = {V{\"o}lske, Michael and Potthast, Martin and Syed, Shahbaz and Stein, Benno},
  booktitle = {Proceedings of the Workshop on New Frontiers in Summarization},
  pages = {59--63},
  year = {2017},
  publisher = {Association for Computational Linguistics}
}

@inproceedings{biderman2023pythia,
  title = {Pythia: A Suite for Analyzing Large Language Models Across Training and Scaling},
  author = {Biderman, Stella and Schoelkopf, Hailey and Anthony, Quentin Gregory and Bradley, Herbie and O'Brien, Kyle and Hallahan, Eric and Khan, Mohammad Aflah and Purohit, Shivanshu and Prashanth, USVSN Sai and Raff, Edward and Skowron, Aviya and Sutawika, Lintang and Van Der Wal, Oskar},
  booktitle = {Proceedings of the 40th International Conference on Machine Learning},
  pages = {2397--2430},
  year = {2023},
  volume = {202},
  series = {Proceedings of Machine Learning Research},
  publisher = {PMLR}
}

@misc{liu2024oat,
  title={OAT: A research-friendly framework for LLM online alignment},
  author={Liu, Zichen and Chen, Changyu and Wan, Xinyi and Du, Chao and Lee, Wee Sun and Lin, Min},
  year={2024},
  howpublished={\url{https://github.com/sail-sg/oat}},
}

\newpage
\appendix
\section{Iteration-wise Win-rate progression}
\label{Iteration-wise-win-rate-progression}
\begin{figure}[htbp]
    \centering
    % First Subfigure
    \begin{subfigure}{0.48\textwidth}
        \centering
        \includegraphics[width=\linewidth]{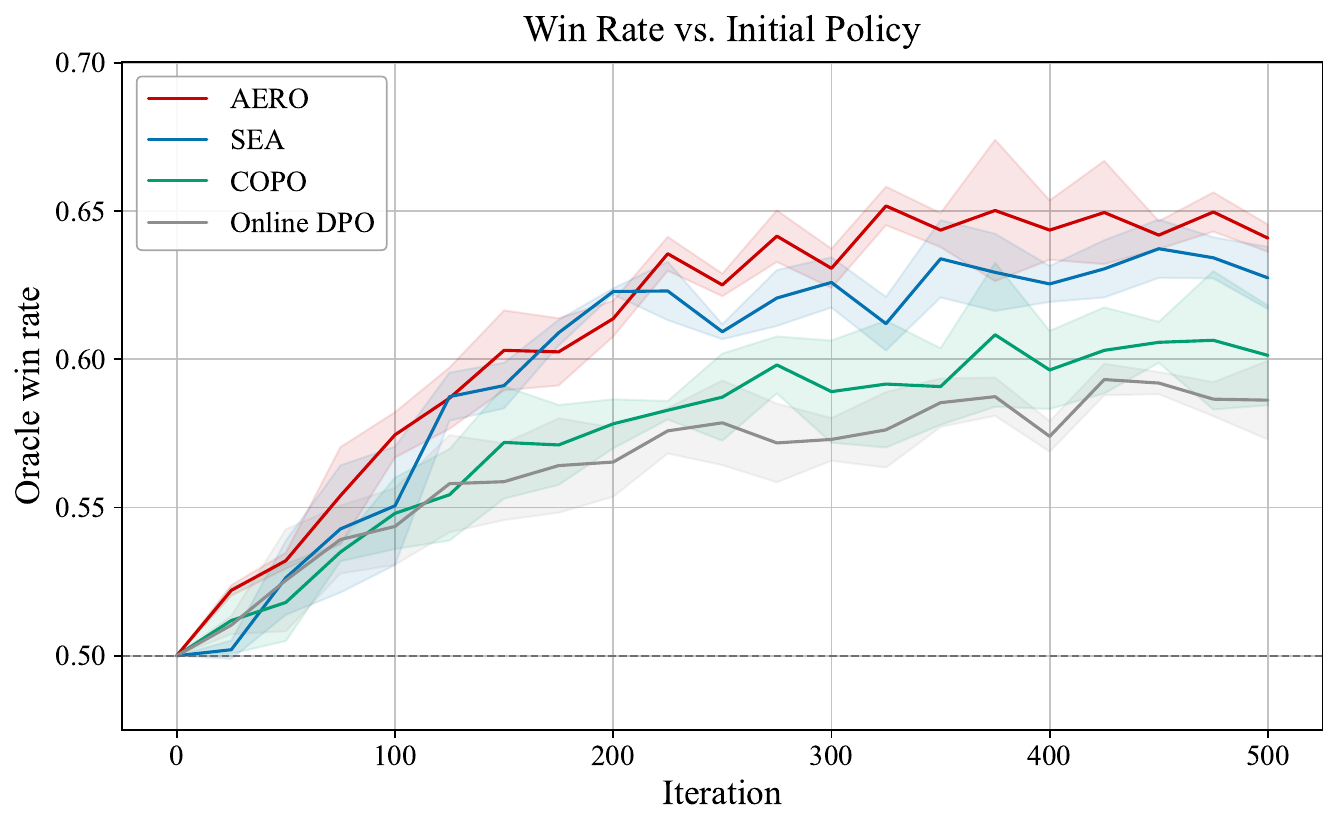}
        \caption{\texttt{Llama-3.2-1B-Instruct}}
        \label{fig:llama_uf_wr_iteration}
    \end{subfigure}\hfill
    % Second Subfigure
    \begin{subfigure}{0.48\textwidth}
        \centering
        \includegraphics[width=\linewidth]{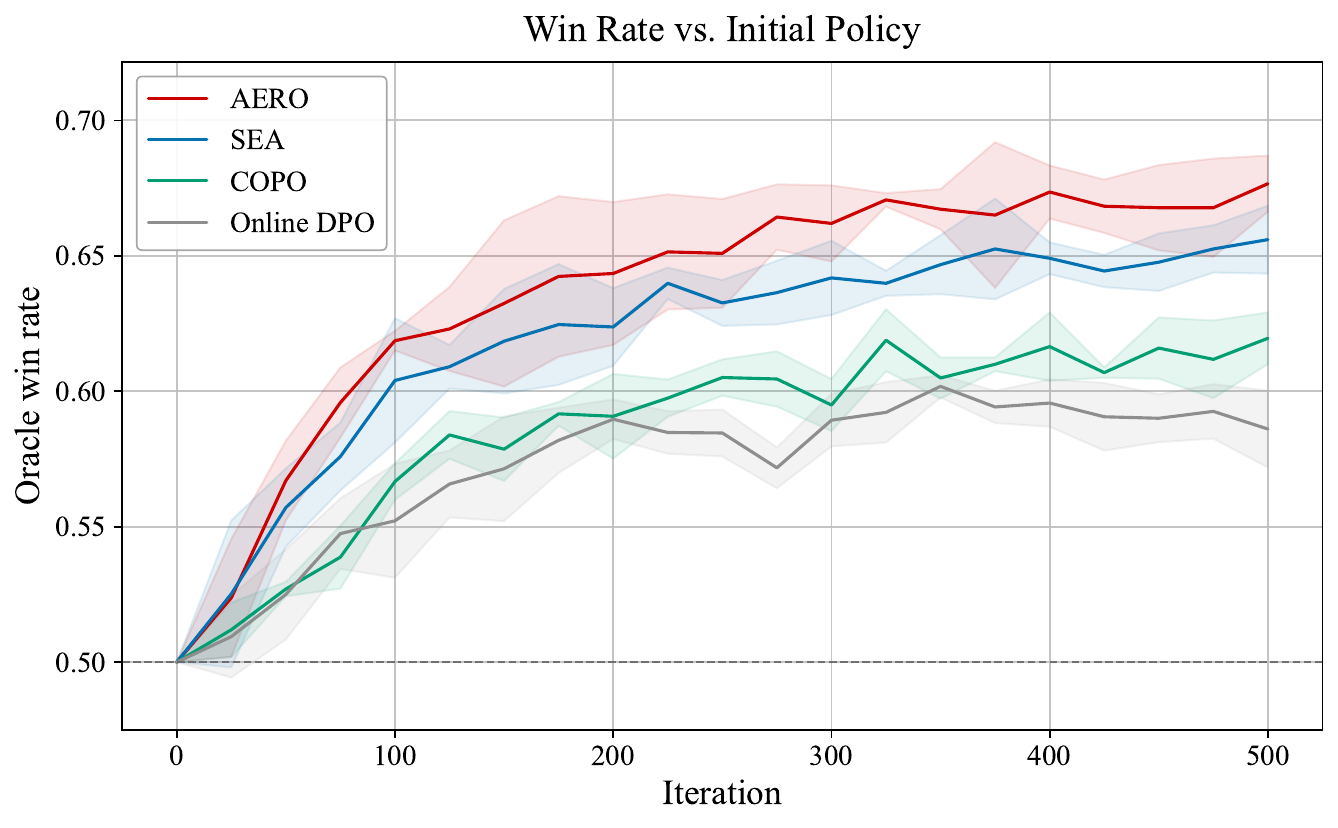}
        \caption{\texttt{Qwen2.5-1.5B-Instruct}}
        \label{fig:qwen15b_uf_wr_iteration}
    \end{subfigure}
    
    \caption{Iteration-wise win rate against the initial instruction-tuned policy. Checkpoints are evaluated every 25 online training iterations.}
    \label{fig:main_results_combined_iteration}
\end{figure}

To complement the oracle-budget plots in Section~\ref{sec:sample_efficiency}, we additionally report iteration-wise win-rate curves. These plots compare methods after the same number of online iterations, rather than only looking at the number of oracle queries. This is to test whether AERO's gains are simply due to routing more samples to ERM synthetic labeling. Across model families, AERO remains competitive or superior through much of the training trajectory, showing that its oracle-budget gains are accompanied by more effective routing of preference supervision and stronger policy improvement per online iteration.

\section{Motivation for Aleatoric Filtering}
\label{Motivation for Aleatoric Filtering}

Section~\ref{sec:pair selectionn} uses the mean entropy of ensemble-head preference predictions as a pair-level proxy for aleatoric ambiguity. Here we motivate this choice through the standard epistemic/aleatoric decomposition. While vote entropy and BAI-style challenger selection serve as practical proxies for epistemic uncertainty, neither is derived from an explicit decomposition between epistemic and aleatoric uncertainties. Following BALD~\citep{houlsby2011bayesian}, epistemic uncertainty can be expressed as the mutual information between predictions and model parameters:
\begin{equation}
    \mathcal{I}(y, \theta | x, \mathcal{D}) = \mathcal{H}(y | x, \mathcal{D}) - \mathbb{E}_{p(\theta | \mathcal{D})}[\mathcal{H}(y | x, \theta)],
\label{bald-eq}
\end{equation}

where $\mathcal{H}(p) = -p \ log \ p - (1-p) \ log \ (1-p)$. Under an ensemble posterior approximation, this corresponds to the entropy of the ensemble-averaged prediction minus the average entropy of individual heads~\citep{kirsch2019batchbald, melo2024deep}.  Intuitively, the first term captures total predictive uncertainty while the second captures ambiguity that persists under a fixed model; their difference therefore isolates uncertainty arising from disagreement across plausible models. Motivated by this decomposition and related entropy-based uncertainty decompositions~\citep{depeweg2018decomposition}, AERO treats the second term as the signal for aleatoric uncertainty. For completeness we restate Eq.~\ref{6}:
\[
    A (x, a, b) = \frac{1}{K}\sum^K_{k=1}\mathcal{H}(p_k(a \succ b \ | \ x))
\]
Large values of \(A(x,a,b)\) occur when the individual heads assign high-entropy
preference probabilities, so the ambiguity is present within the heads themselves
rather than arising only from disagreement between confident heads. To interpret
the threshold, note that binary entropy is symmetric around \(p=0.5\) and maximized
at \(\mathcal{H}(0.5)=\log 2 \approx 0.693\). Solving:
\[
-p\log p-(1-p)\log(1-p)=0.6
\]
gives \(p\approx 0.29\) or \(p\approx 0.71\). Thus, \(A(x,a,b)>0.6\) filters pairs for
which the typical head assigns the comparison a near-indifferent preference probability. Thus, pairs above this threshold are near-indifferent and are filtered out. This is motivated by prior work on uncertainty decomposition, which characterizes aleatoric uncertainty as irreducible noise or ambiguity in the data and epistemic uncertainty as reducible model uncertainty. In Active Learning (AL)~\citep{cohn1996active}, prioritizing epistemic over predictive uncertainty is especially important in noisy preference datasets, where high total uncertainty may be driven by high aleatoric but low epistemic uncertainty~\citep{depeweg2018decomposition, der2009aleatory, kendall2017uncertainties, melo2024deep}. 

\section{Experimental Analysis}
\label{experimental-analysis}
\subsection{Aleatoric Filtering Dynamics}
\begin{figure}[t]
  \centering
  \includegraphics[width=0.80\linewidth]{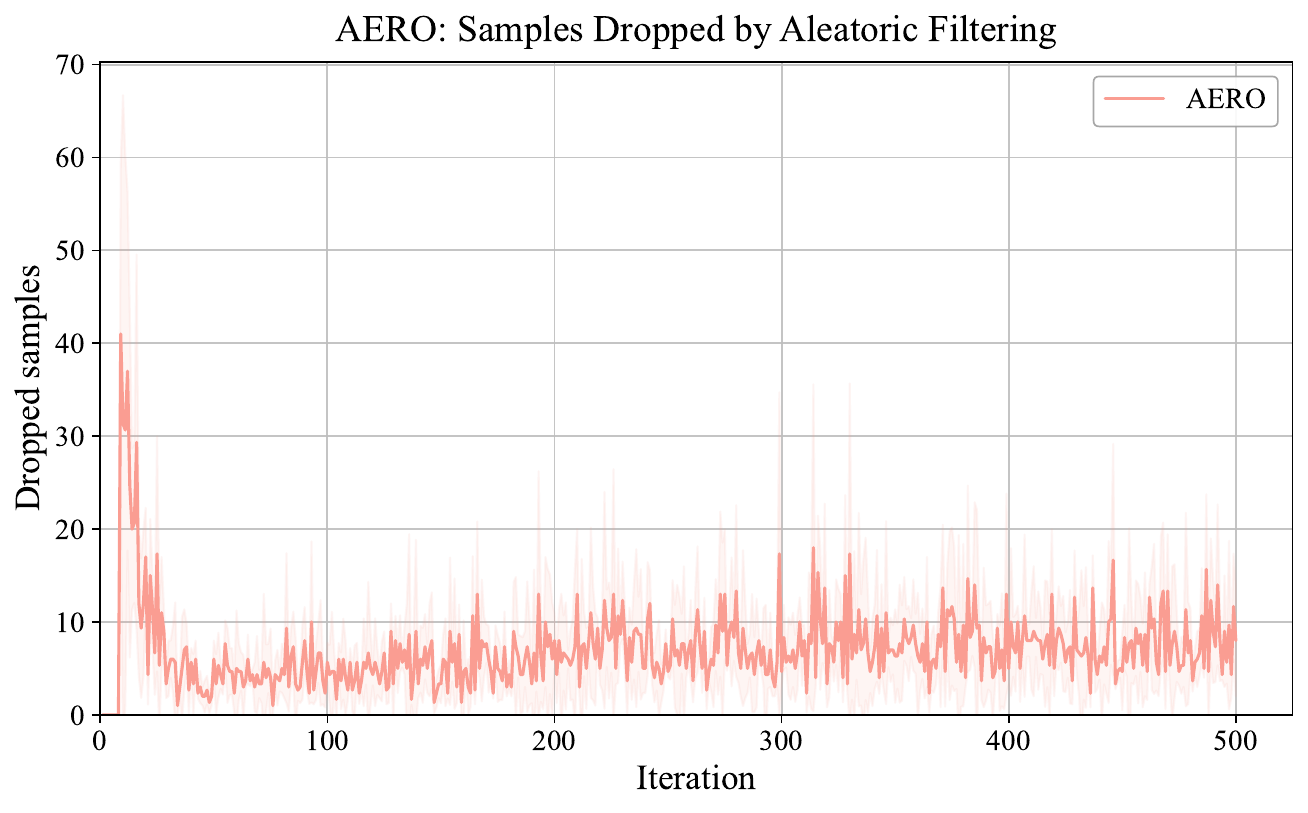}
  \caption{\textbf{Aleatoric filtering activity across training.} Number of candidate pairs dropped per iteration by the aleatoric filter (Eq~\ref{6}, $\alpha$ = 0.6) using \texttt{Llama-3.2-1B-Instruct} on UltraFeedback, averaged across three seeds. The first 8 iterations are a burn-in period during which the filter is disabled.}
  \label{fig:aleatoric_graph}
\end{figure}

AERO uses an 8-iteration burn-in during which the aleatoric filter, adaptive routing, and ERM synthetic labeling are disabled. The ensemble heads are randomly initialized and require some oracle-labeled data before their preference probabilities are calibrated enough to support uncertainty-driven decisions. Both the policy and the ERM therefore receive 8 iterations of unfiltered oracle supervision before AERO's full machinery is utilized.

Figure~\ref{fig:aleatoric_graph} shows the resulting dynamics of the aleatoric filtering. Immediately after the burn-in period, the filter drops a significantly higher number of training samples compared to later iterations. This finding is expected as filtering is enabled only once the ERM has received a small initial amount of oracle supervision, and at that point many candidate comparisons remain low-margin under individual heads, with predicted preference probabilities sitting in the near-indifference band $p \in [0.29, 0.71]$ that corresponds to $A (x, a, b) > 0.6$. As more oracle labels accumulate, the heads sharpen on separable comparisons and the policy distribution adapts, after which the curve settles by roughly iteration 30 into a stable regime of 5–10 dropped examples per iteration. 

The reduction in dropped samples after the post-burn-in spike can also be explained by AERO's use of mixed-preference policy update in Eq.~\ref{7}. SEA~\citep{liu2024sample} uses a TS-based mechanism in which synthetic preferences and pair selection are tied to randomly sampled ensemble heads, thereby keeping the policy coupled to the reward ensemble during online learning. AERO follows the same broad principle of using ERM-generated preferences to shape the on-policy proposal distribution, but follows a more conservative mechanism of using ensemble mean preference probability for both pair selection and labeling, conditioned on low-vote-entropy contexts where the ensemble already exhibits agreement. Thus, AERO does not use synthetic labels to preserve head-specific diversity directly; instead, it uses them to keep the policy synchronized with the ERM's consensus regions. This matters for aleatoric filtering because a context is dropped only when no admissible pair remains. As the policy learns from high-confidence ERM-labeled comparisons, future on-policy candidate sets are more likely to contain at least one comparison that the ensemble can separate confidently, reducing the number of contexts whose candidate pairs are all near-indifferent. Thus, the stabilization after the post-burn-in spike is consistent with ERM-labeled policy updates improving the separability of the generated candidate set $\mathcal{C}_t(x_t)$, while the Variant 3 result in Section~\ref{sec:ablation} suggests that AERO's ensemble-mean based labeling and pair-selection are more stable choices than preserving head-specific pseudo-label diversity in this routed setting.

Quantitatively, AERO trains on $\approx$ 5.5\% fewer preference samples compared to the non-aleatoric filter Variant 1 in Section~\ref{sec:ablation}. This suggests that aleatoric filtering results in modest sample efficiency gains while still maintaining a small but consistent advantage over the Variant 1 ablation. 

\subsection{Adaptive Routing and Uncertainty Diagnostics.}
\paragraph{Vote-entropy distribution and Otsu separability.}
\label{app:vote-entropy-distribution}
We examine whether the batch-level vote-entropy distributions provide a
meaningful partition for Otsu-based adaptive routing. Otsu's criterion selects the threshold from a global property of the score distribution rather than a valley between modes, and therefore does not require the scores to be bimodal~\citep{otsu1979threshold}. Following the Otsu's criterion, we
measure class separability as
$\eta=\sigma_B^2/\sigma_T^2$, where $\sigma_B^2$ and $\sigma_T^2$ denote
between-group and total variance, respectively. Across 1,476 post-burn-in
batches over three seeds, $\eta$ averaged $0.741 \pm 0.035$ and never fell
below $0.633$. In a separate 100-iteration audit run retaining all 128
context-level scores for each of 92 post-burn-in batches, the average
10th, 25th, 50th, 75th, and 90th percentiles of vote entropy were
$0.000$, $0.065$, $0.411$, $0.770$, and $1.040$, respectively. In
particular, the 10th percentile of zero indicates that at least 10\% of
contexts receive unanimous ensemble votes, consistent with the motivation
for routing low-disagreement contexts away from oracle supervision.

We additionally test the shape of these distributions directly using
Hartigan's dip test. All 92 audited post-burn-in batches reject unimodality
at the nominal 5\% level, with a median dip statistic of $0.0630$, median
$p$-value of $0.00046$, and maximum $p$-value of $0.0472$. After Holm
correction across the 92 tests, 80 of 92 batches remain significant. These
results complement Otsu separability statistic by demonstrating that the
observed vote-entropy distributions produced by Otsu-based routing are generally non-unimodal.

\paragraph{Routing behaviour and ensemble diversity.}
Across the 1,476 post-burn-in batches, the fraction of contexts routed to the oracle
before aleatoric filtering ranges from $0.188$ to $0.664$, and no batch is
routed entirely to either branch. Furthermore, Otsu routed a higher fraction of contexts to the oracle in batches with greater overall ensemble-disagreement, as batches in the highest entropy decile have an average
oracle fraction of $0.441$, compared with $0.318$ for those in the lowest
decile. The ensemble remains diverse throughout training, with a mean vote
entropy of $0.444$, $0.508$, and $0.485$ during early, middle, and late
training, respectively.

A potential concern with our adaptive thresholding mechanism arises in edge cases where vote-entropy scores within a batch are uniformly low or high, or show little separation. Although AERO does not impose an absolute entropy floor or ceiling, an oracle-budget cap, or a separability-based fallback, two design choices mitigate such cases in practice. First, routing is disabled during the
eight-iteration burn-in, allowing the ERM to learn from 1,024 oracle-labeled contexts before its disagreement estimates are used for adaptive routing. Second, the ERM uses 20 independently
initialized heads, each regularized toward its own initial weights to preserve diversity in the ensemble. Therefore, to assess whether ensemble diversity degrades over training, we split the post-burn-in iterations into early (9–172), middle (173–336), and late (337–500) phases. Mean vote entropy across the three seeds was 0.444 ± 0.029, 0.508 ± 0.042, and 0.485 ± 0.038, respectively. A collapse in ensemble-diversity would drive vote entropy toward zero, and no such trend appears, indicating that the ensemble remains diverse throughout training. Empirically, we observed no degenerate routing behaviour across the 1,476 post-burn-in batches examined above.

\paragraph{Branch-level uncertainty separation.} To verify that Otsu-based adaptive thresholding produces meaningful separation between the two branches, we log two complementary uncertainty metrics during training. First, we log the average vote entropy (Eq~\ref{3}) for the batch of contexts routed to both the oracle and the ERM. Another metric is the mutual-information-based epistemic uncertainty derived from the decomposition in Eq~\ref{bald-eq}, computed over the selected pair for each branch as $\mathcal{H}(\overline{p}_{ERM}(a \succ b \ | \ x) - \frac{1}{K}\sum^K_{k=1}\mathcal{H}(p_k(a \succ b \ | \ x))$. The two metrics are conceptually distinct: vote entropy measures committee disagreement for a context over candidate completions, while BALD isolates the epistemic component of pairwise preference uncertainty for the chosen pair used for labeling. 

Averaged across three seeds, contexts routed to oracle exhibit roughly 6x higher mean vote entropy than ERM routed contexts, confirming that Otsu's threshold meaningfully partitions each batch into high- and low-disagreement groups rather than producing a near-uniform split. The BALD measurements over labeled pairs show the same direction with a smaller but consistent ratio of approximately 2x. %The absolute BALD values are small in magnitude, which is expected under our shared-backbone ensemble where mutual information is constrained to a narrow range; the relevant quantity is the relative separation between branches.%
The agreement between the two metrics indicates that vote entropy faithfully tracks the epistemic uncertainty that BALD measures at the pair level. This supports our use of vote entropy as the routing signal as oracle queries are concentrated on contexts whose selected pairs carry meaningfully more epistemic uncertainty than those routed to ERM synthetic labeling.

\paragraph{Synthetic-label reliability.}
From our 100-iteration audit run, we also evaluated the reliability of synthetic labels. After the eight-iteration burn-in, we  queried the oracle on every pair that AERO routed to ERM supervision and that passed aleatoric filtering. These audit labels were used only for evaluation. Across 6,381 ERM-routed pairs, the ERM label matched the oracle on 4,876, resulting in an agreement rate of 76.4\%, with mean per-batch agreement of $76.3\% \pm 6.4\%$ over the 92 post-burn-in batches. Low disagreement therefore selects pairs whose synthetic labels agree with the oracle in a large majority of cases, but does not guarantee similarity in preference; we treat the residual disagreement as synthetic-label noise. Importantly, the ERM trains only on oracle labels, so this noise enters the policy update but is never fed back into the ensemble.

\subsection{Compute and Resource Accounting}
\label{app:compute_accounting}

Oracle-query efficiency does not account for the additional candidate-generation
and ERM cost incurred by AERO and the active exploration baselines. We therefore additionally report performance in terms of
compute-efficiency and total resource usage for the primary
three-seed \texttt{Llama-3.2-1B-Instruct} experiment on UltraFeedback. GPU-hours are
estimated from measured runtimes, and wall-clock time is reported per seed
using two A100 40GB GPUs.

\begin{figure}[t]
    \centering
    \includegraphics[width=0.80\linewidth]{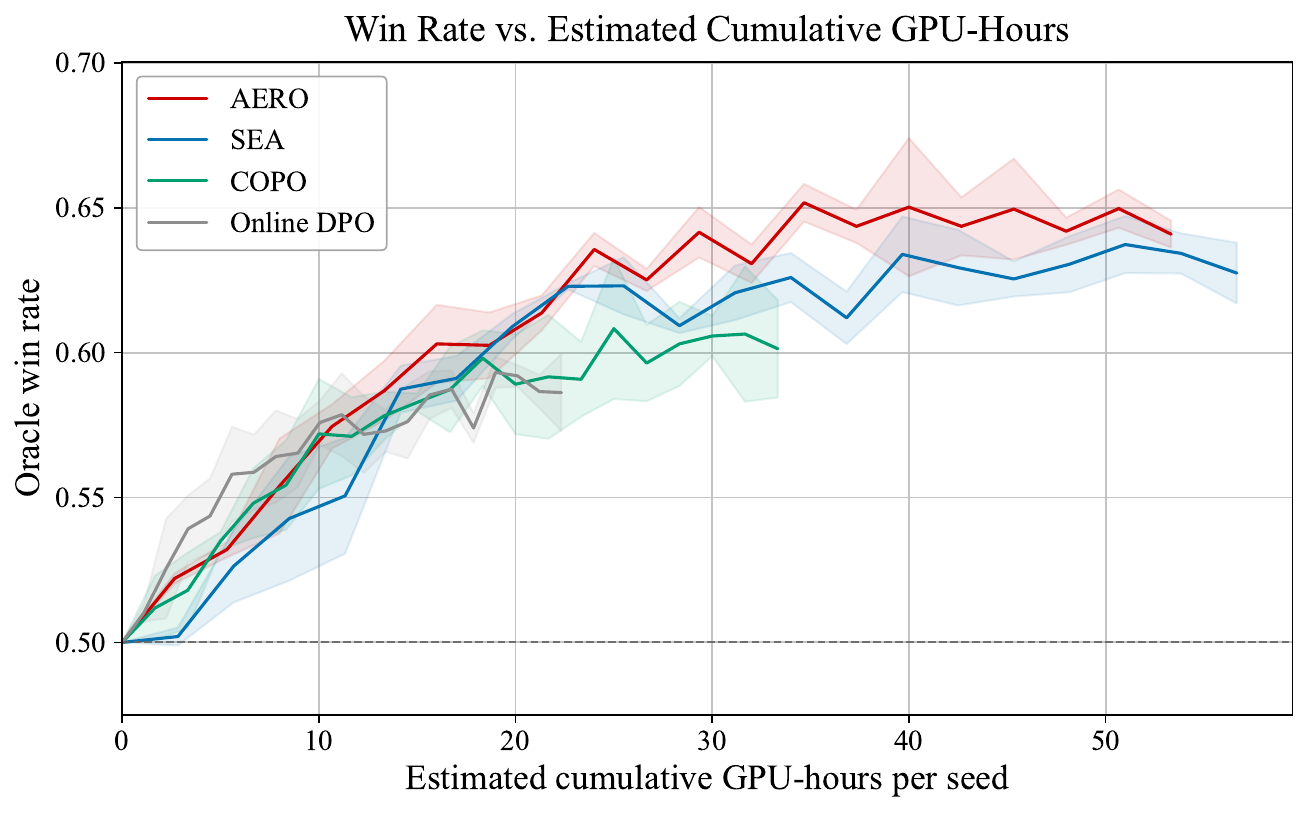}
    \caption{Oracle win rate against estimated cumulative GPU-hours per seed
    on \texttt{Llama-3.2-1B-Instruct} and UltraFeedback.}
    \label{fig:compute_normalized}
\end{figure}

AERO and SEA have broadly the same dominant additional compute cost induced by generating
$N=20$ responses per context and adopting a
20-head ERM, which gives 3.84M candidate generations across three seeds.
The 20 heads operate on each shared encoding rather than requiring 20 separate
backbone passes. AERO's vote entropy, Otsu thresholding, and pair filtering
reuse these scores and add no additional model forward passes.

\begin{table}[t]
\centering
\caption{Total resource usage for the primary three-seed
Llama-3.2-1B-Instruct experiment. Generation, erm-routing, synthetic-label,
oracle-query, and GPU-hour counts are totals across three seeds, whereas wall-clock
time is per seed on two A100 40GB GPUs.}
\vspace{6pt}
\label{tab:resource_accounting}
\resizebox{\linewidth}{!}{
\begin{tabular}{lrrrrrr}
\toprule
\textbf{Method} &
\textbf{Generations} &
\textbf{ERM-routed} &
\textbf{Synthetic labels} &
\textbf{Oracle queries} &
\textbf{GPU-h} &
\textbf{Wall-clock} \\
\midrule
AERO       & 3.840M & 102.0K & 99.7K & 81.8K  & $\sim$160 & $\sim$27 h \\
SEA        & 3.840M & 56.1K  & 56.1K & 135.9K & $\sim$170 & $\sim$28 h \\
COPO       & 0.576M & 0      & 0     & 192.0K & $\sim$100 & $\sim$17 h \\
Online DPO & 0.384M & 0      & 0     & 192.0K & $\sim$67  & $\sim$11 h \\
\bottomrule
\end{tabular}}
\end{table}

\Cref{fig:compute_normalized} shows that against Online DPO and COPO, the trade-off can be seen at lower compute budgets,
as Online DPO initially achieves higher win rates for the compute
expended, while COPO largely tracks AERO. However, AERO overtakes both at approximately 12 GPU-hours
per seed and subsequently achieves higher win rates at comparable cumulative
compute budgets. Against SEA, which has an identical compute profile by
construction, AERO achieves higher compute-normalized performance for most of
training, with the advantage becoming more significant later in training.
Overall, although AERO requires greater learner compute than Online DPO and
COPO, this additional computation translates into both substantially lower
oracle usage and stronger compute-normalized performance as training progresses.

\subsection{Cross-Evaluator Analysis}
\label{app:evaluator_independence}

Our main evaluation uses the \texttt{Skywork-Reward-Llama-3.1-8B} RM, which also provides the oracle
preference labels. We therefore compare its gains with those from
length-controlled AlpacaEval 2.0 within each training stage. Specifically,
we compute the cross-method Spearman rank agreement over the four methods at
the early, middle, and final checkpoints.

\begin{table}[t]
\centering
\caption{Within-stage Spearman agreement between win-rate gains evaluated using the \texttt{Skywork-Reward-Llama-3.1-8B} RM and AlpacaEval 2.0 length-controlled win-rate gains.}
\vspace{6pt}

\label{tab:evaluator_agreement}
\begin{tabular}{lccc}
\toprule
\textbf{Model} & \textbf{Early} & \textbf{Middle} & \textbf{Final} \\
\midrule
Llama-3.2-1B & 0.60 & 1.00 & 0.40 \\
Qwen2.5-1.5B & 0.40 & 1.00 & 1.00 \\
Llama-3-8B   & 0.80 & 0.60 & 0.80 \\
\bottomrule
\end{tabular}
\end{table}

As shown in~\Cref{tab:evaluator_agreement}, the agreement is positive in all nine model-by-stage comparisons, although
each correlation comparison contains only four methods and should therefore be interpreted
cautiously. The two evaluations also differ substantially, as for evaluation using the \texttt{Skywork-Reward-Llama-3.1-8B} RM, we use
held-out UltraFeedback prompts that are not used for training. At each evaluated
checkpoint, the current policy and the initial instruction-tuned policy each
generate one response to the same prompt, and the RM selects the
preferred response; the reported win-rate is the fraction of comparisons won
by the current policy. AlpacaEval 2.0 instead evaluates responses on 805
general instruction-following prompts from the AlpacaFarm evaluation set,
comparing each model response against a fixed GPT-4-Turbo reference response
using a GPT-4-Turbo auto-annotator. These
correlations use matched single-run checkpoints and the three-seed endpoint
estimates in \Cref{tab:alpaca_three_models} are not used here.

\paragraph{Arena-Hard.}
We also evaluate the final \texttt{Llama-3-8B-Instruct} checkpoints on the 500 Arena-Hard prompts
using responses generated by the canonical GPT-4-0314 and GPT-4.1 as the judge.

\begin{table}[t]
\centering
\caption{Arena-Hard results for the final Llama-3-8B checkpoints.}
\label{tab:arena_hard}
\begin{tabular}{lcc}
\toprule
\textbf{Method} & \textbf{Raw WR} & \textbf{Style-controlled WR} \\
\midrule
Initial policy & 18.0 & 17.0 ($-1.2/+1.7$) \\
Online DPO     & 18.2 & 17.7 ($-1.2/+1.3$) \\
SEA            & 16.6 & 15.8 ($-1.5/+1.3$) \\
COPO           & 15.4 & 15.1 ($-1.5/+1.2$) \\
AERO           & 19.0 & 17.2 ($-1.4/+1.8$) \\
\bottomrule
\end{tabular}
\end{table}

As shown in ~\Cref{tab:arena_hard}, AERO has the highest Raw WR, but after Arena-Hard's official
length-and-markdown style control, the confidence intervals overlap substantially
across methods. In particular, Arena-Hard does not clearly separate AERO,
Online DPO, and the initial policy, while the initial policy also ranks above
two aligned baselines. This suggests limited separation among these particular
8B checkpoints. We therefore treat Arena-Hard as a complementary evaluation
and retain length-controlled AlpacaEval 2.0 as our primary external evaluation.
Furthermore, because GPT-4.1 differs from the original Arena-Hard judge, the Arena-Hard scores only serve as within-study comparisons.

\subsection{Additional Ablations}

\label{app:fixed_quantile}

\begin{figure}[t]
    \centering
    \includegraphics[width=0.80\linewidth]{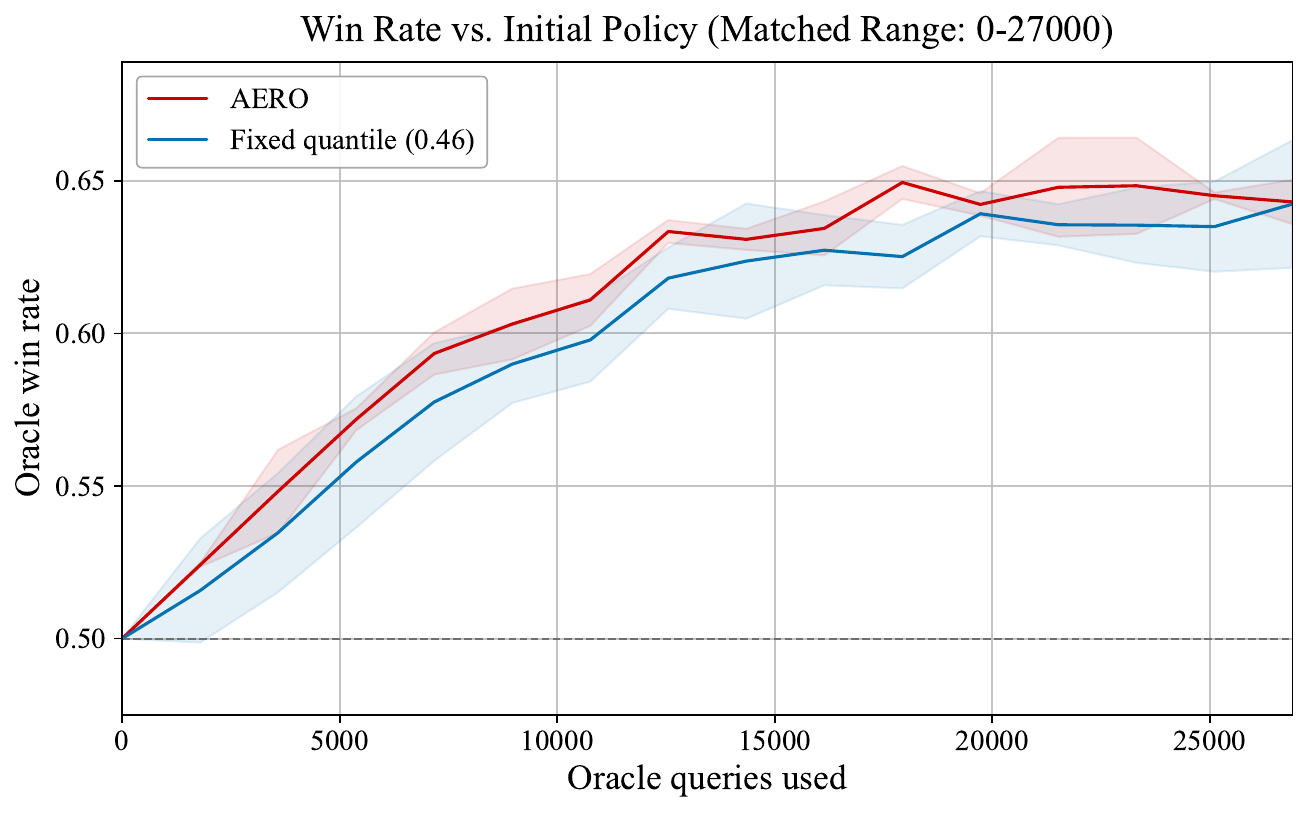}
    \caption{\textbf{Adaptive versus fixed-quantile routing.} Comparison of AERO’s adaptive Otsu routing with fixed-quantile routing using an oracle-routing ratio of \(q=0.46\), evaluated on \texttt{Llama-3.2-1B-Instruct} with UltraFeedback and averaged across three seeds.}
    \label{fig:aero_vs_fixed_quantile}
\end{figure}

\paragraph{Comparison with fixed-quantile routing.} We compare original AERO against a fixed-quantile routing ablation to investigate whether the benefits of Otsu-based adaptive routing extend beyond removing the need to tune a routing ratio. The fixed quantile router retains vote-entropy based routing and all other components but replaces Otsu's adaptive threshold with a fixed top-$q$ vote-entropy rule at AERO's average pre-aleatoric-filter allocation on UltraFeedback ($q = 0.46$) (\Cref{fig:aero_vs_fixed_quantile}). AERO achieves a slightly higher mean win rate across most of the matched oracle budget range, and its matched-budget area under the win-rate curve is higher in two of three paired runs, whereas final win rates are effectively identical. Unlike the random-routing comparison (\Cref{fig:aero_vs_variant2_27k}), this advantage is not consistent across seeds, so we interpret it as a modest average benefit of batch-adaptive routing rather than a reliable improvement. The comparison mainly indicates that, once contexts are routed using vote entropy, a well-chosen fixed fraction recovers most of the adaptive-routing benefit. However, the ratio of $0.46$ was itself obtained automatically using Otsu that adapts between settings ($0.460$ on UltraFeedback versus $0.322$ on TL;DR), while a fixed-quantile router would otherwise require a per-setting hyperparameter search.

\begin{table}[t]
\centering
\caption{Single-seed sweep over fixed routing fractions $q$ on \texttt{Llama-3.2-1B-Instruct} with UltraFeedback. As in \Cref{fig:main_results_combined}, WR (\%) is the held-out win rate against the initial policy judged by \texttt{Skywork-Reward-Llama-3.1-8B}, and $B$ is the cumulative oracle queries. For AERO (Otsu, average fraction $\approx 0.46$) we report the range over three seeds. Per-iteration WR differs by only 1--2 points with no consistent ordering across $q$, while $B$ scales with $q$; confirming this trend requires additional seeds.}
\vspace{6pt}
\label{tab:quantile_sweep}
\setlength{\aboverulesep}{0pt}
\setlength{\belowrulesep}{0pt}
\renewcommand{\arraystretch}{1}
\resizebox{\linewidth}{!}{%
\begin{tabular}{c cc|cc|cc|cc}
\toprule
\multirow{2}{*}{\textbf{Iteration}}
& \multicolumn{2}{c|}{\textbf{Fixed $q = 0.15$}}
& \multicolumn{2}{c|}{\textbf{Fixed $q = 0.25$}}
& \multicolumn{2}{c|}{\textbf{AERO (Otsu)}}
& \multicolumn{2}{c}{\textbf{Fixed $q = 0.65$}} \\
\cmidrule(lr){2-3}
\cmidrule(lr){4-5}
\cmidrule(lr){6-7}
\cmidrule(lr){8-9}
& \textbf{WR} & $B$
& \textbf{WR} & $B$
& \textbf{WR} & $B$
& \textbf{WR} & $B$ \\
\midrule
100 & 57.84 & 2.8K  & 59.88 & 3.9K  & 56.67--58.20 & 5.2--5.7K   & 58.91 & 8.4K \\
200 & 64.71 & 4.7K  & 62.47 & 7.1K  & 61.00--62.07 & 10.8--11.4K & 63.24 & 16.4K \\
300 & 63.65 & 6.6K  & 64.10 & 10.3K & 62.42--63.75 & 16.1--16.9K & 63.59 & 24.2K \\
400 & 64.77 & 8.5K  & 65.43 & 13.5K & 63.29--65.27 & 21.6--22.2K & 64.56 & 32.1K \\
500 & 66.19 & 10.4K & 65.58 & 16.7K & 63.75--64.61 & 26.9--27.6K & 66.80 & 39.8K \\
\bottomrule
\end{tabular}%
}
\end{table}

\paragraph{Hyperparameter sweep over quantile values.} We perform a modest single-seed sweep over fixed quantile values $q \in \{0.15, 0.25, 0.65\}$ (\Cref{tab:quantile_sweep}) to investigate whether disagreement-based routing performs well at lower oracle ratios, and to examine the trade-off against an automatic thresholding method such as Otsu. Our results suggest that even at oracle ratios as low as $0.15$, the remaining components of AERO retain the performance achieved at the ratio selected by Otsu, with per-iteration win rates comparable to AERO and differences likely within run-to-run or seed variation. Nonetheless, these results reveal an important limitation of Otsu as even a modest sweep identifies fractions that are substantially more oracle-efficient than the one selected by Otsu. However, such a sweep must be repeated whenever the model, dataset, or model size changes, which is computationally expensive. Since Otsu is a statistical rather than a decision-theoretic criterion, it provides a pragmatic, tuning-free way to select the oracle fraction, albeit a conservative one in terms of oracle savings. This trade-off motivated our use of Otsu in AERO's design. Finally, this sweep is only evaluated using the \texttt{Skywork-Reward-Llama-3.1-8B} oracle, which also provides the training labels. AlpacaEval 2.0 is a stronger evaluation, as it uses an independent and more capable GPT-4-Turbo judge and its LC win rate correlates strongly with human preference rankings; hence, whether the oracle-efficiency gains of lower fractions transfer to it remains to be verified.

\section{Reproducibility details}
\label{appendix: reproducibility}

\begin{figure}[t]
  \centering
  \includegraphics[width=0.80\linewidth]{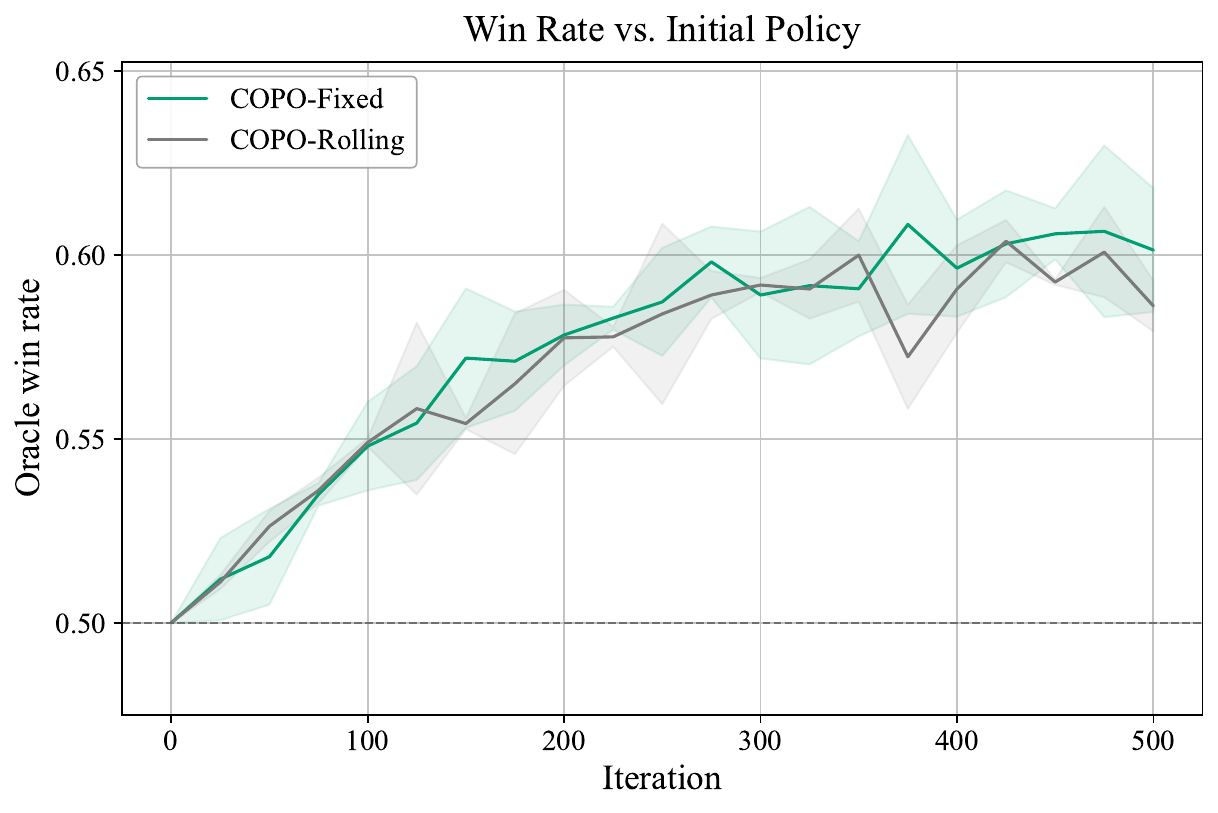}
  \caption{Comparison of fixed-CFN and rolling-CFN variants of the COPO-adapted baseline on \texttt{Llama-3.2-1B-Instruct} with UltraFeedback. The fixed-CFN version is used in the main results because it performs slightly better in our fine-grained online setting.}
  \label{fig:copo_vs_copo2}
\end{figure}

We include COPO \citep{bai2025online} as a representative baseline for optimism-based online preference alignment. The original COPO pipeline operates over a small number of large online rounds. In each round, one response is generated from the current policy for each prompt, ranked together with the dataset's existing chosen and rejected responses using PairRM~\citep{jiang2023llm}, and the best/worst pair is used for DPO training. COPO additionally trains a Coin Flip Network (CFN) to estimate a pseudo-count exploration bonus, following the original CFN method \citep{lobel2023flipping}.

Our setting differs from the original COPO method because all methods are evaluated under a fully on-policy pairwise-preference oracle budget. We therefore implement a COPO-adapted baseline rather than an exact reproduction. First, we discard static dataset responses and sample fresh responses from the current policy, matching the rest of our online baselines. Second, we do not use the oracle to rank all three generated candidates, since a three-way ranking doesn't match the cumulative budget $B$ we use for other methods which is based on the pairwise label provided by an oracle for a DAP policy training objective. Instead, we use a learned ensemble as a low-cost internal scorer to select a candidate pair from three on-policy responses, and then query the oracle once for the final label before policy training. This is to ensure that the COPO-adapted baseline uses the same pairwise oracle budget as the other methods.

The CFN schedule is the main implementation choice (\Cref{fig:copo_vs_copo2}). We considered two variants:
\begin{itemize}
    \item \textbf{COPO-fixed}: the CFN is trained once on 5k initial prompt-response pairs, where each response is generated by the initial policy, and then kept fixed throughout training. In this variant, the CFN acts as a stable novelty prior for the count-based DPO bonus.

    \item \textbf{COPO-rolling}: as a more dynamic CFN variant, motivated by the original CFN method \citet{lobel2023flipping}, we update the CFN online from a rolling buffer of recent prompt-response pairs, where the responses are generated by the current policy. In the original CFN algorithm, visited states are stored in a replay buffer and the CFN is updated during training to produce count-based exploration bonuses; our rolling variant adapts this idea to prompt-response pairs in online preference optimization. After each three-response candidate set is constructed, the remaining third response is used as the COPO reference response and added to the CFN buffer before the DPO step.
\end{itemize}

\section{Implementation Details}
\label{app:implementation}
\begin{table}[htbp]
    \centering
    \label{tab:hyperparameters}
    
    \begin{tabular}{l >{\raggedright\arraybackslash}p{9cm}}
        \toprule
        \multicolumn{2}{c}{\textbf{Basic Parameters}}\\
        \midrule
        Fine-tuning strategy & LoRA~\citep{hu2022lora} (\texttt{Llama-3-8B-Instruct}), Full parameter fine tuning (\texttt{Qwen2.5-1.5B-Instruct}, \texttt{Llama-3.2-1B-Instruct}) \\
        LoRA target modules & q\_proj, k\_proj, v\_proj, o\_proj, gate\_proj, up\_proj, down\_proj \\
        LoRA $r$ & 64 \\
        LoRA alpha & 64 \\
        LoRA dropout & 0.05 \\
        Learning rate & $5 \times 10^{-7}$ (\texttt{Qwen2.5-1.5B-Instruct}, \texttt{Llama-3.2-1B-Instruct}), $2 \times 10^{-5}$ (\texttt{Llama-3-8B-Instruct})  \\
        Learning rate warmup ratio & 0.02 (\texttt{Qwen2.5-1.5B-Instruct}, \texttt{Llama-3.2-1B-Instruct}), 0.005 (\texttt{Llama-3-8B-Instruct}) \\
        Learning rate scheduler & cosine \\
        Train batch size & 128 \\
        Training epochs & 1 \\
        DPO $\beta$ & 0.1 \\
        Optimizer & Adam \\
        ERM backbone & llm-blender/PairRM \\
        Ensemble size & 20 \\
        Number of on-policy responses & 20 (AERO, SEA), 3 (COPO), 2 (DPO) \\
        vLLM generation temperature & 0.7 (during training), greedy decoding for evaluation \\
        Reward Oracle & \texttt{Skywork-Reward-Llama-3.1-8B}~\citep{liu2024skywork} \\
        ERM regularization coefficient $\lambda$ & 0.5 \\
        Training Iterations & 500 (\texttt{Qwen2.5-1.5B-Instruct}, \texttt{Llama-3.2-1B-Instruct}), 300 (\texttt{Llama-3-8B-Instruct}) \\
        \bottomrule
    \end{tabular}
\end{table}

Parts of our reward-modeling implementation and SEA baseline are adapted from the OAT framework~\citep{liu2024oat}. Our ERM follows SEA~\citep{liu2024sample} as we remove the head of a pretrained PairRM~\citep{jiang2023llm} encoder and add an ensemble of $K=20$ MLP reward heads, each with two hidden layers of size 128. Following SEA and ~\citet{dwaracherla2024efficient}, the ERM is intentionally much smaller than the reward oracle. For reproducing SEA, we use the hyperparameters recommended in their paper. For COPO, we keep the published hyperparameter configuration fixed, including the CFN architecture, training schedule, regularization, and optimizer settings, and tune only the exploration factor $\alpha$ within our adapted fully on-policy implementation.

For \texttt{Llama-3-8B-Instruct}, we use LoRA rather than full-parameter fine-tuning for memory and compute efficiency. Because LoRA freezes the base model and updates only low-rank adapter parameters, its optimization dynamics are not directly comparable to the full-parameter 1B and 1.5B model runs. The learning rate used for full-parameter fine-tuning runs ($5 \times 10^{-7}$) produced little measurable improvement, so we tuned the LoRA learning rate and selected $2\times10^{-5}$ as the most stable setting. We use a shorter warmup ratio and stop at 300 iterations because the 8B LoRA runs converged and plateaued earlier than the 1B full-finetuning runs, making longer training computationally unnecessary.

\paragraph{Reward Oracle.} Following SEA~\citep{liu2024sample}, we use \texttt{Skywork-Reward-Llama-3.1-8B} \citep{liu2024skywork} as the preference oracle across all experiments. The oracle is a deterministic scalar reward model and does not capture the stochasticity of real human preference judgments. We adopt it for consistency with prior work in this line and to isolate algorithmic differences from oracle-noise effects.

\paragraph{Alpaca Evaluation.} For AlpacaEval 2.0, we generate one response per instruction using vLLM with temperature $0.7$ and top-$p=1.0$ for all methods. We use identical decoding parameters across methods and report AlpacaEval 2.0 only for controlled within-paper comparisons rather than leaderboard comparison. The original AlpacaEval 2.0 GPT-4-Turbo evaluator uses \texttt{gpt-4-1106-preview}. Since \texttt{gpt-4-1106-preview} is no longer available through the OpenAI API, we pin the evaluator to \texttt{gpt-4-turbo-2024-04-09}, a later GPT-4-Turbo snapshot, for reproducibility. 

\paragraph{Hardware Details.} All \texttt{Llama-3.2-1B-Instruct} experiments were conducted on two A100 40GB GPUs. The \texttt{Qwen2.5-1.5B-Instruct} and \texttt{Llama-3-8B-Instruct} experiments used a g7e.24xlarge instance on AWS with four NVIDIA RTX 6000 Pro 96GB GPUs available. Under our compute-credit constraints, we allocated two GPUs for each \texttt{Llama-3-8B-Instruct} run and one GPU for each \texttt{Qwen2.5-1.5B} run. 

\section{Additional Pythia-TL;DR Results}
\label{app:pythia_tldr}

\begin{figure}[h!]
    \centering
    \begin{subfigure}{0.49\linewidth}
        \centering
        \includegraphics[width=\linewidth]{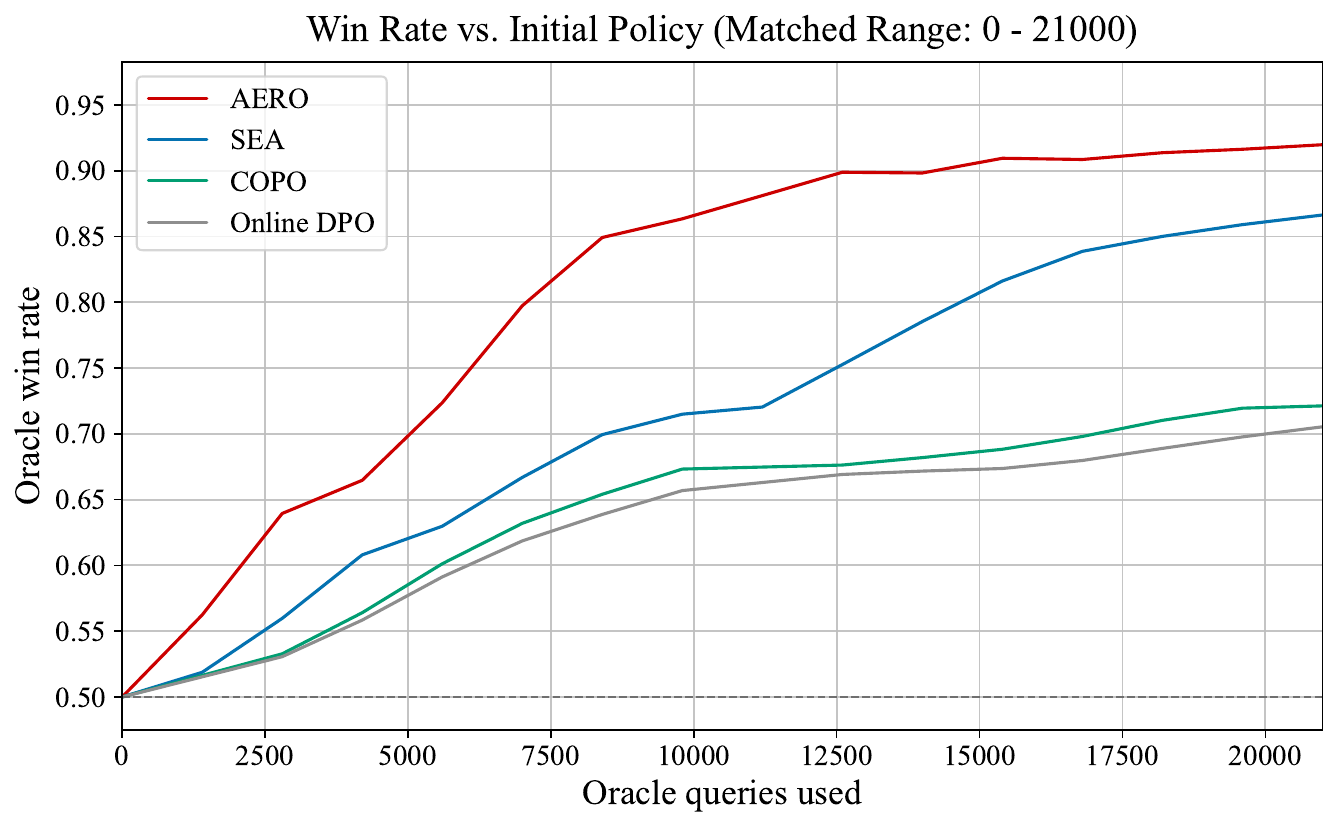}
        \caption{Win rate by oracle budget.}
    \end{subfigure}
    \hfill
    \begin{subfigure}{0.49\linewidth}
        \centering
        \includegraphics[width=\linewidth]{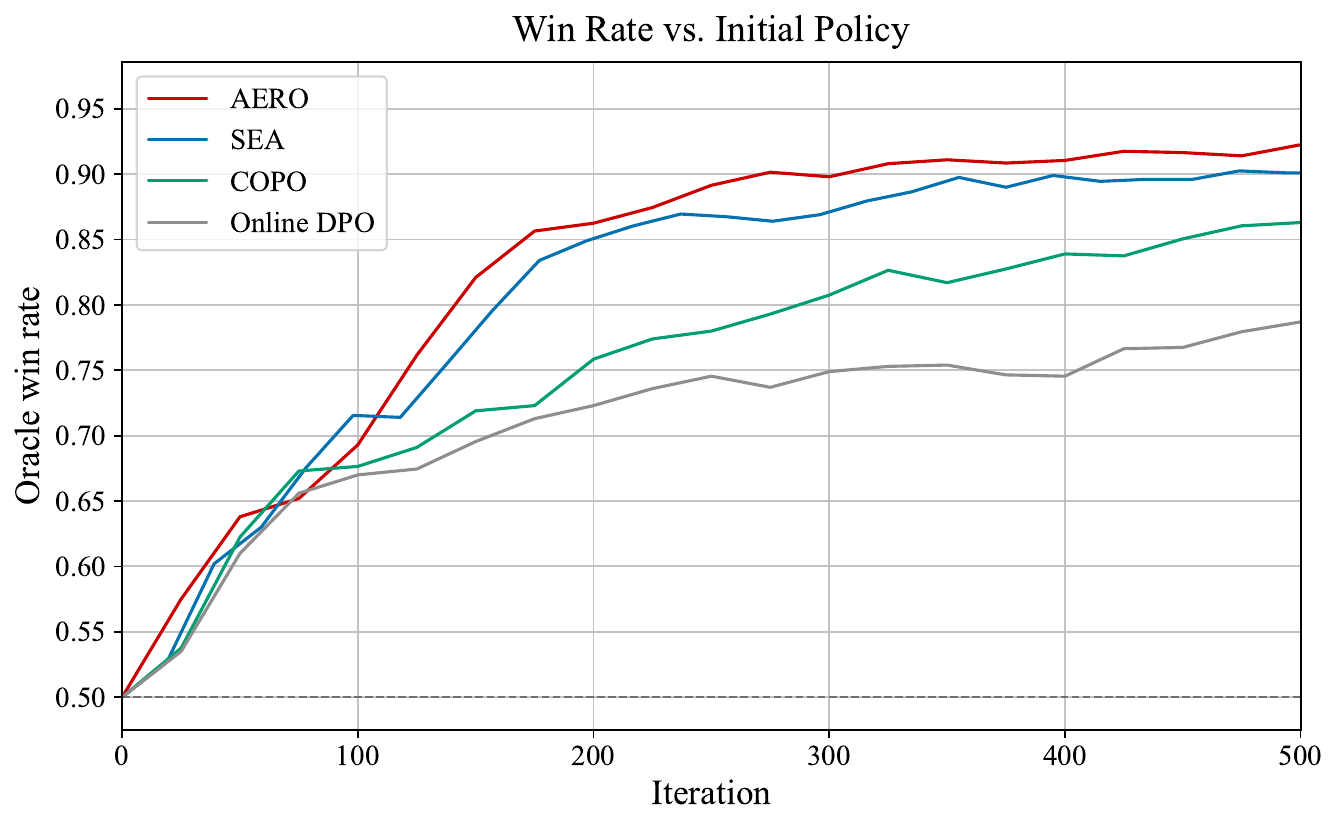}
        \caption{Win rate by iteration.}
    \end{subfigure}
    \caption{Single-seed results on TL;DR summarization~\citep{volske2017tldr} using a 1B model from the Pythia family~\citep{biderman2023pythia}. AERO performs comparably to SEA by iteration and achieves a higher win rate at matched oracle budgets. The oracle-budget plot shows that Otsu-based routing selects a smaller oracle budget in this easier summarization setting, suggesting that AERO adapts by routing more low-disagreement contexts to ERM synthetic labeling when the ensemble reaches consensus.}
    \label{fig:pythia_tldr}
\end{figure}
\newpage
\end{document}